\documentclass[10pt,twocolumn,letterpaper]{article}

\usepackage{cvpr} %

\usepackage{times}
\usepackage{epsfig}
\usepackage{graphicx}
\usepackage{amsmath}
\usepackage{amssymb}
\usepackage{subfig}

\usepackage{adjustbox}

\usepackage{tabularx}
\usepackage{booktabs} %

\usepackage[numbers,sort]{natbib}
\usepackage[breaklinks=true,bookmarks=false]{hyperref}

\usepackage[margin=4pt,font=footnotesize,labelfont=bf,labelsep=endash,tableposition=top]{caption}

\cvprfinalcopy %

\newcommand\T{\rule{0pt}{1.2ex}}       %
\newcommand\B{\rule[-0.8ex]{0pt}{0pt}} %
\def\etal{\emph{et al}.}

\def\cvprPaperID{5072} %

\begin{document}

\title{Fast Neural Architecture Search of Compact Semantic Segmentation Models \\
via Auxiliary Cells}%

\author{Vladimir Nekrasov\thanks{Equal contribution. }
\quad
Hao Chen\footnotemark[1]
\quad
Chunhua Shen
\quad
Ian Reid \\
The University of Adelaide, Australia\\
{
 \small
 E-mail:
 \{vladimir.nekrasov, hao.chen01, chunhua.shen, ian.reid\}@adelaide.edu.au 
}
}

\maketitle
\begin{abstract}
Automated design of neural network architectures tailored for a specific task is an extremely promising, albeit inherently difficult, avenue to explore. While most results in this domain have been achieved on image classification and language modelling problems, here we concentrate on dense per-pixel tasks, in particular, semantic image segmentation using fully convolutional networks. 
In contrast to the aforementioned areas, the design choices of a fully convolutional network require several changes, ranging from the sort of operations that need to be used---e.g., dilated convolutions---to a solving of a more difficult optimisation problem.
In this work, we 
are particularly interested in
searching for high-performance compact segmentation architectures, able to run in real-time using limited resources. To achieve that, we intentionally over-parameterise the architecture during the training time via a set of auxiliary cells that provide an intermediate supervisory signal and can be omitted during the evaluation phase.  The design of the auxiliary cell is emitted by a controller, a neural network with the fixed structure trained using reinforcement learning. More crucially, we demonstrate how to efficiently search for these architectures within limited time and computational budgets. In particular, we rely on a progressive strategy that terminates non-promising architectures from being further trained, and on Polyak averaging coupled with knowledge distillation to speed-up the convergence. Quantitatively, in $8$ GPU-days our approach discovers a set of architectures performing on-par with state-of-the-art among compact models on the semantic segmentation, pose estimation and depth prediction tasks. %
Code will be made available here: \url{https://github.com/drsleep/nas-segm-pytorch}

\vskip -0.2in
\end{abstract}

\section{Introduction}

For years, the design of neural network architectures was thought to be solely a duty of a human expert - it was her responsibility to specify which type of architecture to use, how many layers should there be, how many channels should convolutional layers have and etc. This is no longer the case as the automated neural architecture search - a way of predicting the neural network structure via a non-human expert (an algorithm) - is fast-growing. Potentially, this may well mean that instead of manually adapting a single state-of-the-art architecture for a new task at hand, the algorithm would discover a set of best-suited and high-performing architectures on given data.

Few decades ago, such an algorithm was based on evolutionary programming strategies where best seen so far architectures underwent mutations and their most promising off-springs were bound to continue evolving~\cite{AngelineSP94}. Now, we have reached the stage where a secondary neural network, oftentimes called \emph{controller}, replaces a human in the loop, by iteratively  searching among possible architecture candidates and maximising the expected score on the held-out set~\cite{ZophL16}. While there is a lack of theoretical work behind this latter approach, several promising empirical breakthroughs have already been achieved~\cite{BakerGNR16,ZophVSL17}.

At this point, it is important to emphasise the fact that such accomplishments required an excessive amount of computational resources---more than $20,000$ GPU-days for the work of Zoph and Le~\cite{ZophL16} and $2,000$ for Zoph~\etal~\cite{ZophVSL17}. Although a few works have reduced those to single digit numbers on image classification and language processing tasks~\cite{PhamGZLD18,LiuZNSHLFYHM18}, we consider more challenging dense per-pixel tasks that produce an output for each pixel in the input image and for which no efficient training regimes have been previously presented. Although here we concentrate only on semantic image segmentation, our proposed methodology can immediately
be applied to other per-pixel prediction tasks, such as depth estimation and
pose estimation. In our experiments, we demonstrate the transferability of the discovered segmentation architecture to the latter problems. Notably, all of them play an important role in computer vision and robotic applications and so far have been relying on manually designed accurate low-latency models for real-world scenarios.

The focus of our work is to automatically discover compact high-performing fully convolutional architectures, able to run in real-time on a low-computational budget, for example, on the Jetson platform. To this end, we are explicitly looking for structures that not only improve the performance on the held-out set, but also facilitate the optimisation during the training stage. Concretely, we consider the encoder-decoder type of a fully-convolutional network~\cite{LongSD15}, where encoder is represented by a pre-trained image classifier, and the decoder structure is emitted by the controller network. The controller generates the connectivity structure between encoder and decoder, as well as the sequence of operations (that form the so-called \emph{cell}) to be applied on each connected path. The same cell structure is used to form an auxiliary classifier, the goal of which is to provide intermediate supervision and to implicitly over-parameterise the model. Over-parameterisation is believed to be the primary reason behind the successes of deep learning models, and a few theoretical works have already addressed it in simplified cases~\cite{SoltanolkotabiJ17,pmlr-v80-du18a}. Along with empirical results, this is the primary motivation behind the described approach.

Last, but not least, we devise a search strategy that permits to find high-performing architectures within a small number of days using only few GPUs. Concretely, we pursue two goals here: 
\begin{itemize}
\itemsep -.122cm
    \item[i.)] 
     To prevent `bad' architectures from being trained for long; and 
\item[ii.)] 
     To achieve a solid performance estimate as soon as possible.
\end{itemize}
To tackle the first goal, we divide the training process during the search into two stages. During the first stage, we fix the encoder's weights and pre-compute its outputs, while only training the decoder part.
For the second stage, we train the whole model end-to-end. We validate the performance after the first stage and terminate the training of non-promising architectures. For the second goal, we employ Polyak averaging~\cite{polyak1992acceleration} and knowledge distillation~\cite{HintonVD15} to speed-up convergence.

To summarise, our contributions in this work are to
      propose an efficient neural architecture search strategy for dense-per-pixel tasks
     that (i.) allows to sample compact high-performing architectures,
     and (ii.) can  be used in real-time on low-computing platforms, such as JetsonTX2.
In particular, the above points are made possible by:
\begin{itemize}
\itemsep -.122cm
    \item Devising a progressive strategy able to eliminate poor candidates early in the training;
    \item Developing a training schedule for semantic segmentation able to provide solid results quickly via the means of knowledge distillation and Polyak averaging;
    \item Searching for an over-parameterised auxiliary cell that provides better training and is obsolete during inference.
\end{itemize}

\section{Related Work}

Traditionally, architecture search methods have been relying upon evolutionary strategies~\cite{AngelineSP94,StanleyM02,StanleyDG09}, where a population of networks (oftentimes together with their weights) is continuously mutated, and less promising networks are being discarded. Modern neuro-evolutionary approaches~\cite{RealMSSSTLK17,abs-1711-00436} rely on the same principles and benefit from available computational resources, that allow them to achieve impressive results. Bayesian optimisation methods estimating the probability density of objective function have long been used for hyper-parameter search~\cite{SnoekLA12,BergstraYC13}. Scaling up Bayesian methods for architecture search is an ongoing work, and few kernel-based approaches have already shown solid performance~\cite{swersky2014raiders,abs-1802-07191}.

Most recently, neural architecture search (NAS) strategies based on reinforcement learning (RL) have attained state-of-the-art results on the tasks of image classification and natural language processing~\cite{BakerGNR16,ZophL16,ZophVSL17}. Relying on enormous computational resources, these algorithms comprise a separate neural network, the so-called {\em `controller'}, that emits an architecture design and receives a scalar reward after the emitted architecture is trained on the task of interest. Notably, thousand of iterations and GPU-days are needed for convergence. Rather than searching for the whole network structure from scratch, these methods tend to look for \emph{cells}---repeatable motifs that can be stacked multiple times in a feedforward fashion.

Several solutions for making NAS methods more efficient have been recently proposed.
In particular, Pham~\etal~\cite{PhamGZLD18} unroll the computational graph of all possible architectures and allow sharing the weights among different architectures. This dramatically reduces the number of resources needed for convergence. In a similar vein of research, Liu~\etal~\cite{LiuZNSHLFYHM18} exploit a progressive strategy where the network complexity is gradually increased, while the ranking network is trained in parallel to predict the performance of a new architecture. A few methods have been built around continuous relaxation of the search problem. Particularly Luo~\etal~\cite{abs-1808-07233} use an encoder to embed the architecture description into a latent space, and estimator to predict the performance of an architecture given its embedding. While these methods make the search process more efficient, they achieve so by sacrificing the expressiveness of the search space, and hence, may arrive to a sub-optimal solution.

In semantic segmentation \cite{Lin2016EfficientPT,LinMSR17,Lin2017Semantic}, up to now all the architectures have been manually designed, closely following the winner entries of image classification challenges. Two prominent directions have emerged over the last few years: the encoder-decoder type~\cite{LongSD15,NohHH15,LinMSR17}, where better features are learned at the expense of having a spatially coarse output mask; whereas other popular approach discards several down-sampling layers and relies on dilated convolutions for keeping the receptive field size intact~\cite{ChenPKMY18,ZhaoSQWJ17,YuKF17}. Chen~\etal~\cite{ChenZPSA18} have also shown that the combination of those two paradigms lead to even better results across different benchmarks. In terms of NAS in semantic segmentation, independently of us and in parallel to our work, a straightforward adaptation of image classification NAS methods was proposed by Chen~\etal~\cite{abs-1809-04184}. In it they randomly search for a single segmentation cell design and achieve expressive results by using {\em almost $400$~GPUs over the range of $7$ days}.  In contrast to that, our method first and foremost is able to find compact segmentation models only in a fraction of that time. Secondly, it differs significantly in terms of the search design and search methodology.

For the purposes of a clearer presentation of our ideas, we briefly review knowledge distillation, an approach proposed by Hinton~\etal~\cite{HintonVD15} to successfully train a compact model using the outputs of a single (or an ensemble of) large network(s) pre-trained on the current task. In it, the logits of the pre-trained network are being used as an additional regulariser for the small network.  In other words, the latter has to mimic the outputs of the former. Such a method was shown to provide a better learning signal for the small network.
As a result of that, it has already found its way across multiple domains: computer vision~\cite{abs-1804-08328}, reinforcement learning~\cite{RusuCGDKPMKH15}, continuous learning~\cite{LiH16} -- to name a few.

\section{Methodology}

We  start with the problem formulation, proceed with the definitions of an auxiliary cell and knowledge distillation loss, and conclude with the overall search strategy.

We primarily focus on two research questions: (i.) how to acquire a reliable estimate of the segmentation model performance as quickly as possible; and (ii.) how to improve the training process of the segmentation architecture through over-parameterisation, obsolete during inference.

\subsection{Problem Formulation}
We consider dense prediction task $T$, for which we have multiple training tuples $\{(X_{i},y_{i})\}$, where both $X_{i}$ and $y_{i}$ are $3$-dimensional tensors with equal spatial and arbitrary third dimensions. In this work, $X_{i}$ is a $3$-channel RGB image, while $y_{i}$ is a $C$-channel one-hot segmentation mask with $C$ being equal to the number of classes, which corresponds to semantic image segmentation. Furthermore, we rely on a mapping $f: X\rightarrow y$ with parameters $\theta$, that is represented by a fully convolutional neural network.  We assume that the network $f$ can further be decomposed into two parts: $e$ - representing encoder, and $d$ - for decoder. We initialise encoder $e$ with weights from a pre-trained classification network consisting of multiple down-sampling operations that reduce the spatial dimensions of the input. The decoder part, on the other hand, has access to several outputs of encoder with varying spatial and channel dimensions. The search goal is to choose which feature maps to use and what operations to apply on them. We next describe the decoder search space in full detail.%

\subsubsection{Search Space}
\label{sec:search-space}
We restrict our attention to the decoder part, as it is currently infeasible to perform a full segmentation network search from scratch. 

\begin{figure*}[t!]
	\centering
	\subfloat{\includegraphics[width = 0.97\linewidth]{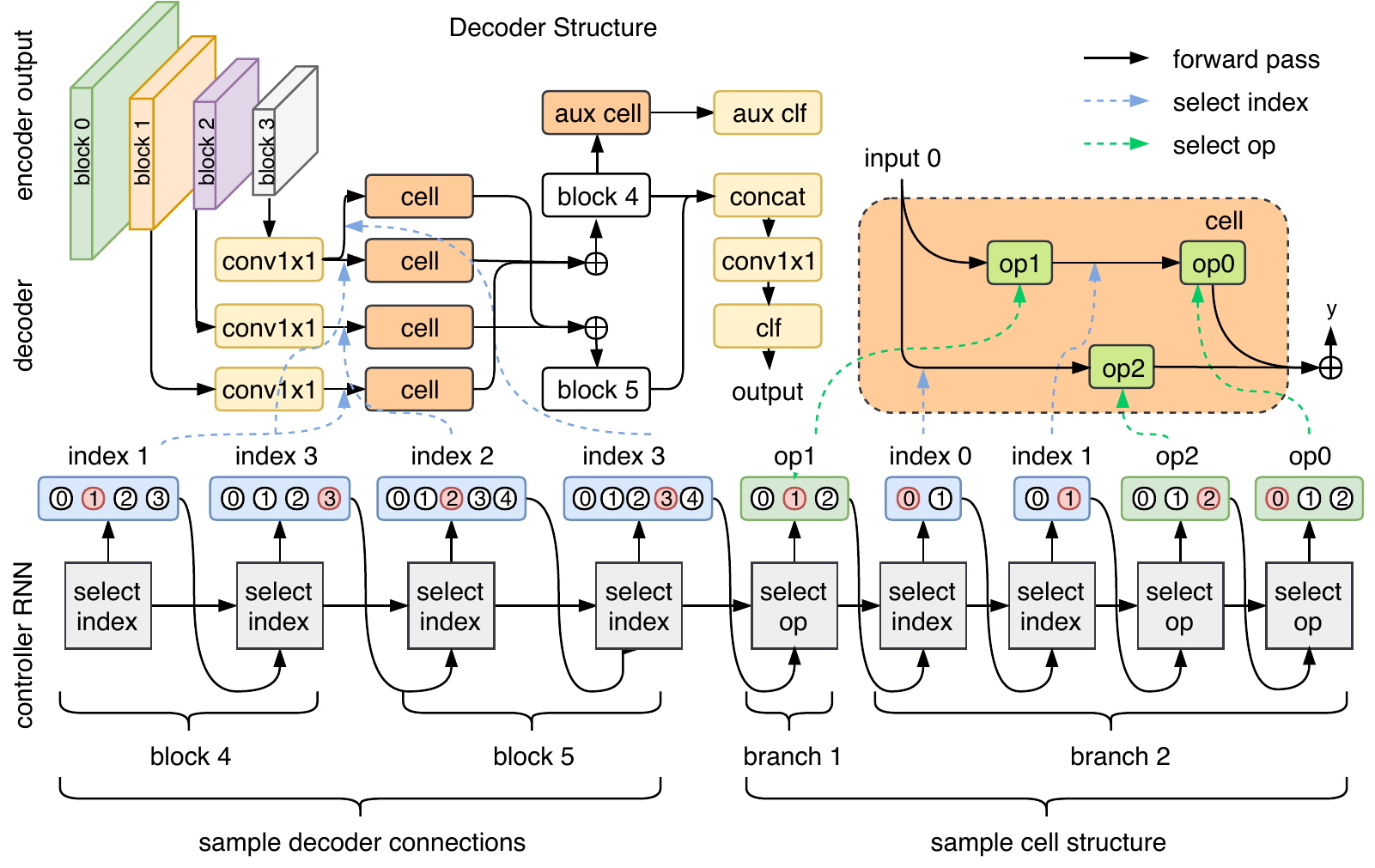}}
	\caption{Example of the encoder-decoder auxiliary search layout. Controller RNN (\emph{bottom}) first generates connections between encoder and decoder (\emph{top left}), and then samples locations and operations to use inside the cell (\emph{top right}). All the cells (including auxiliary cell) share the emitted design.\\
	In this example, the controller first samples two indices (\emph{block1} and \emph{block3}), both of which pass through the corresponding cells, before being summed up to create \emph{block4}. The controller then samples \emph{block2} and \emph{block3} that are merged into \emph{block5}. Since \emph{block4} was not sampled, it is concatenated with \emph{block5} and fed into $1$$\times$$1$ convolution followed by the final classifier. The output of \emph{block4} is also passed through an auxiliary cell for intermediate supervision. To emit the cell design, the controller starts by sampling the first operation applied on the cell input (\emph{op1}), followed by sampling of two indices -- \emph{index0}, corresponding to the cell input, and \emph{index1} of the output layer after the first operation. Two operations -- \emph{op2} and \emph{op0} -- are applied on each index, respectively, and their summation serves as the cell output.\label{fig:search-aux}}
\end{figure*}

As mentioned above, the decoder has access to multiple layers from the pre-trained encoder with varying dimensions. To keep sampled architectures compact and of approximately equal size, each encoder output undergoes a single $1$$\times$$1$ convolution with the same number of output channels. We rely on a recurrent neural network, the controller, to sequentially produce pairs of indices of which layers to use, and what operations to apply on them. In particular, this sequence of operations is combined to form a cell (see example in Fig.~\ref{fig:search-aux}). The same cell but with different weights is applied to each layer inside the sampled pair, and the outputs of two cells are summed up. The resultant layer is added to the sampling pool. The number of times pairs of layers are sampled is controlled by a hyper-parameter, which we set to $3$ in our experiments, allowing the controller to recover such encoder-decoder architectures as FCN~\cite{LongSD15}, or RefineNet~\cite{LinMSR17}. All non-sampled summation outputs are concatenated, before being fed into a single $1\times1$ convolution to reduce the number of channels followed by the final classification layer.

Each cell takes a single input with the controller first deciding which operation to use on that input. The controller then proceeds by sampling with replacement two locations out of two,  i.e., of input and the result of the first operation, and two corresponding operations. The outputs of each operation are summed up, and all three layers (from each operation and the result of their summation) together with the initial two can be sampled on the next step. The number of times the locations are sampled inside the cell is controlled by another hyper-parameter, which we also set to $3$ in our experiments in order to keep the number of all possible architectures to a feasible amount\footnote{Taking into account symmetrical -- thus, identical -- architectures, we estimate the number of unique connections in the decoder part to be $120$, and the number of unique cells $\sim$$10^{10}$, leading to $\sim$$10^{12}$, which is on-par with concurrent works.}. All existing non-sampled summation outputs inside the cell are summed up, and used as the cell output. In this case, we resort to sum as concatenation may lead to variable-sized outputs between different architectures.

Based on existing research in semantic segmentation, we consider $11$ operations:
\begin{itemize}
\itemsep -.122cm
\item conv $1\times1$,
\item conv $3\times3$,
\item separable conv $3\times3$,
\item separable conv $5\times5$,
\item global average pooling followed by upsampling and conv $1\times1$,
\item conv $3\times3$ with dilation rate $3$,
\item conv $3\times3$ with dilation rate $12$,
\item separable conv $3\times3$ with dilation rate $3$,
\item separable conv $5\times5$ with dilation rate $6$,
\item skip-connection,
\item zero-operation that effectively nullifies the path.
\end{itemize}
An example of the search layout with $2$ decoder blocks and $2$ cell branches is depicted on Fig.~\ref{fig:search-aux}.\footnote{Please refer to {\em Appendix A} for more details on the search space and the sampling procedure.}

\subsection{Search Strategy}
We randomly divide the training set into two disjoint sets - meta-train and meta-val. The meta-train subset is used to train the sampled architecture on the given task (i.e., semantic segmentation), whereas meta-val, on the other hand, is used to evaluate the trained architecture and provide the controller with a scalar, oftentimes called \emph{reward} in the reinforcement learning literature. Given the sampled sequence, its logarithmic probabilities and the reward signal, the controller is optimised via proximal policy optimisation (PPO)~\cite{schulman2017proximal}. Hence, there are two training processes present: inner - optimisation of the sampled architecture on the given task, and outer - optimisation of the controller. We next concentrate on the inner loop.

\subsubsection{Progressive Stages}
We divide the inner training process into two stages. During the first stage, the encoder weights are fixed and its outputs are pre-computed, while only decoder is being trained. This leads to a quick adaptation of the decoder weights and a reasonable estimate of the performance of the sampled architecture. We exploit a simple heuristic to decide whether to continue training the sampled architecture for the second stage, or not. Concretely, the current reward value is being compared with the running mean of rewards seen so far, and if it is higher, we continue training. Otherwise, with probability $1-p$ we terminate the training process. The probability $p$ is annealed throughout our search (starting from $0.9$).

The motivation behind this is straightforward: the results of the first stage, while noisy, can still provide a reasonable estimate of the potential of the sampled architecture. At the very least, they would present a reliable signal that the sampled architecture is non-promising, while spending only few seconds on it. Such a simple approach encourages exploration during early stages of search akin to the $\epsilon$-greedy strategy often used in the multi-armed bandit problem~\cite{watkins1989learning}.

\subsubsection{Fast Training via Knowledge Distillation and Weights' Averaging}

Semantic segmentation models are notable for requiring many iterations to converge. Partially, this is addressed by initialising the encoder part from a pre-trained classification network. Unfortunately, no such thing exists for decoder.

Fortunately, though, we can explore several alternatives that provide faster convergence. Besides tailoring our optimisation hyper-parameters, we rely on two more tricks: firstly, we keep track of the running average of the parameters during each stage and apply them before the final validation~\cite{polyak1992acceleration}. %
Secondly, we append an additional $l_{2}-$loss term between the logits of the current architecture and a pre-trained teacher network. We can either pre-compute the teacher's outputs beforehand, or acquire them on-the-fly in case the teacher's computations are negligible.

The combination of both of these approaches allows us to receive a very reliable estimate of the performance of the semantic segmentation model as quickly as possible without a significant overhead.

\subsubsection{Intermediate Supervision via Auxiliary Cells}

We further look for ways of easing optimisation during fast search, as well as during a longer training of semantic segmentation models. %
Thus, still aligning with the goal of having a compact but accurate model, we explicitly aim to find ways of performing steps that are beneficial during training and obsolete during evaluation.

One approach that we consider here is to append an auxiliary cell after each summation between pairs of main cells - the auxiliary cell is identical to the main cell and can either be conditioned to output ground truth directly, or to mimic the teacher's network predictions (or the combination of the above two). At the same time, it does not influence the output of the main classifier either during the training or testing and merely provides better gradients for the rest of the network.
In the end, the reward per the sampled architecture will still be decided by the output of the main classifier. For simplicity, we only apply the segmentation loss on all auxiliary outputs.

The notion of intermediate supervision is not novel in neural networks, but to the best of our knowledge, prior works have merely been relying on a simple auxiliary classifier, and we are the first to tie up the design of decoder with the design of the auxiliary cell. We demonstrate the quantitative benefits of doing so in our ablation studies (Sect.~\ref{subsec-eff}).

Furthermore, our motivation behind searching for cells that may also serve as intermediate supervisors stems from ever-growing empirical (and theoretical under certain assumptions) evidence that deep networks benefit from over-parameterisation during training~\cite{SoltanolkotabiJ17,pmlr-v80-du18a}. %
While auxiliary cells provide an implicit notion of over-parameterisation, we could have explicitly increased the number of channels and then resorted to pruning. Nonetheless, pruning methods tend to result in unstructured networks often carrying no tangible benefits in terms of the runtime speed, whereas our solution simply permits omitting unused layers during inference. %

\section{Experiments}

We conduct extensive experiments on PASCAL VOC which is  an established semantic segmentation benchmark that comprises $20$ semantic classes (and background) and provides $1464$ training images~\cite{EveringhamGWWZ10}. For the search process, we extend those to more than $10000$ by exploiting annotations from BSD~\cite{HariharanABMM11}. As commonly done, during search, we keep $10\%$ of those images for validation of the sampled architectures that provides the controller with the reward signal. For the first stage, we pre-compute the encoder outputs on $4000$ images and store them for faster processing.

The controller is a two-layer recurrent LSTM~\cite{HochreiterS97} neural network with $100$ hidden units. All the units are randomly initialised from a uniform distribution. We use PPO~\cite{schulman2017proximal} for optimisation with the learning rate of $0.0001$. 

The encoder part of our network is MobileNet-v2~\cite{abs-1801-04381}, pretrained on MS COCO~\cite{LinMBHPRDZ14} for semantic segmentation using the Light-Weight RefineNet decoder~\cite{NekrasovS018}. We omit the last layers and consider four outputs from layers $2,3,6,8$ as inputs to decoder; $1$$\times$$1$ convolutional layers used for adaptation of the encoder outputs have $48$ output channels during search and $64$ during training. Decoder weights are randomly initialised using the Xavier scheme~\cite{GlorotB10}. To perform knowledge distillation, we use Light-Weight RefineNet-152~\cite{NekrasovS018}, and apply $\ell_{2}-$loss with the coefficient of $0.3$ which was set using the grid search. The knowledge distillation outputs are pre-computed for the first stage and omitted during the second one in the interests of time. Polyak averaging is applied with the decay rates of $0.9$ and $0.99$, correspondingly. Batch normalisation statistics are updated during both stages.

All our search experiments are being conducted on two $1080$Ti GPU cards, with the search process being terminated after $4$ days. All runtime measurements are carried out on a single $1080$Ti card, or on JetsonTX2, if mentioned otherwise. In particular, we perform the forward pass $100$ times and report the mean result together with standard deviation. 

\subsection{Search Results}

\begin{figure}[t!]
	\centering
	\subfloat{\includegraphics[width = 0.99\linewidth, clip, trim=0 0 0 5]{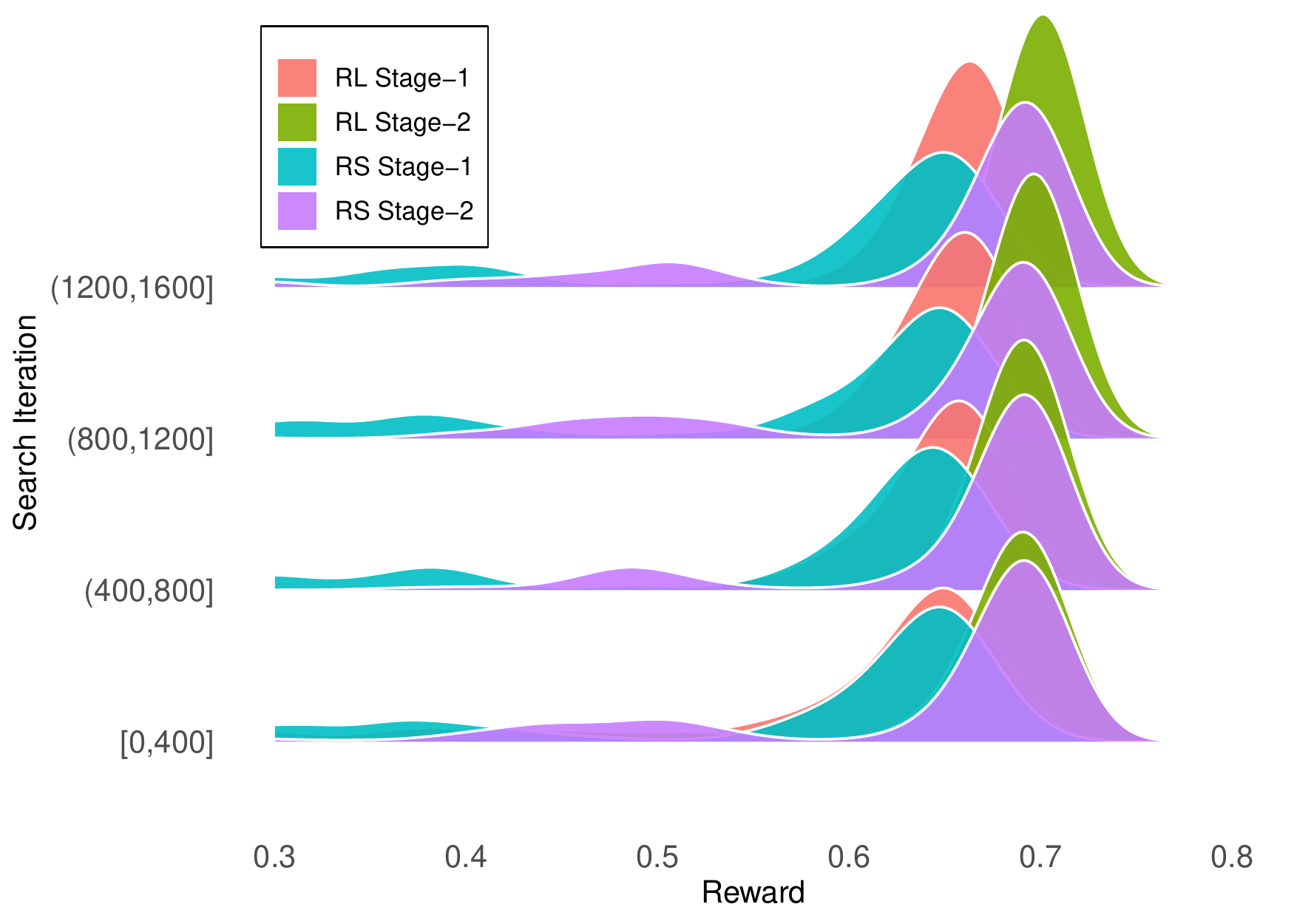}}
	\vskip -0.1in
	\caption{Distribution of rewards per each training stage for reinforcement learning (\emph{RL}) and random search (\emph{RS}) strategies. Higher peaks correspond to higher density.
	\label{fig:r-t01}}
	\vskip -0.15in
\end{figure}

\begin{figure}[b!]
	\centering
	\includegraphics[width=0.99\linewidth, clip, trim=0 20 0 0]{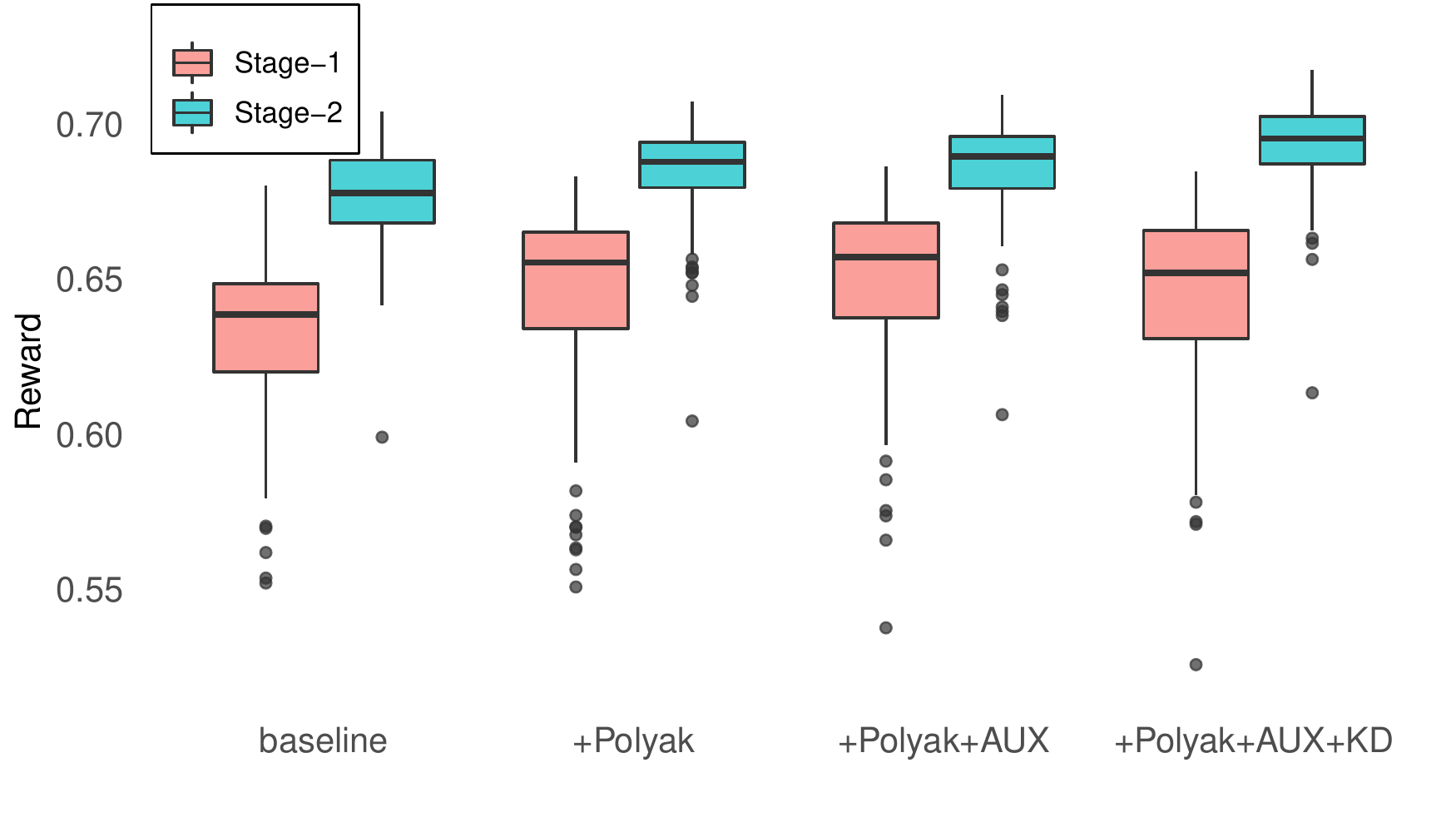}
	\vskip -0.1in
	\caption{Distribution of rewards during each training stage of the search process across setups with Polyak averaging (\emph{Polyak}), intermediate supervision through auxiliary cells (\emph{AUX}) and knowledge distillation (\emph{KD}). 
	}
	\label{fig:pol-aux-kd}
\end{figure}

For the inner training of the sampled architectures, we devise a fast and stable training strategy: we exploit the Adam learning rule~\cite{KingmaB14} for the decoder part of the network, and SGD with momentum - for encoder. In particular, we use learning rates of $3e$-$3$ and $1e$-$3$, respectively. We pre-train each sampled architecture for $5$ epochs on the first stage, and for $1$ on the second (in case the stopping criterion is not triggered). As the reward signal, we consider the geometric mean of three quantities: namely, 

i.) mean intersection-over-union (IoU), or Jaccard Index~\cite{EveringhamGWWZ10}, primarily used across semantic segmentation benchmarks;

ii.) frequency-weighted IoU, that scales each class IoU by the number of pixels present in that class, and

iii.) mean-pixel accuracy, that averages the number of correct pixels per each class. When computing, we do not include background class as it tends to skew the results due to a large number of pixels belonging to background. As mentioned above, we keep the running mean of rewards after the first stage to decide whether to continue training a sampled architecture.

We visualise the reward progress during both stages on Figure~\ref{fig:r-t01}. As evident from it, the quality of the emitted architectures grows with time - it is even possible that more iterations would lead to better results, although we do not explore that to save the time spent. On the other hand, while random search has the potential of occasionally sampling decent architectures, it finds only a fraction of them in comparison to the RL-based controller.

Moreover, we evaluate the impact of the inclusion of Polyak averaging, auxiliary cells and knowledge distillation on each training stage. To this end, we randomly sample and train $140$ architectures. We visualise the distributions of rewards on Fig.~\ref{fig:pol-aux-kd}. All the tested settings significantly outperform baseline on both stages, and the highest rewards on the second stage are attained when using all of the components above. 

\subsection{Effect of Intermediate Supervision via Auxiliary Cells}
\label{subsec-eff}

After the search process is finished, we select $10$ architectures discovered by the RL controller with highest rewards and proceed by carrying out additional ablation studies aimed to estimate the benefit of the proposed auxiliary scheme in case the architectures are allowed to train for longer.

In particular, we train each architecture for $20$ epochs on BSD together with PASCAL VOC and $30$ epochs on PASCAL VOC only. For simplicity, we omit Polyak averaging and knowledge distillation. Three distinct setups are being tested: concretely, we estimate whether intermediate supervision helps at all, and whether auxiliary cell is superior to a plain auxiliar classifier

\begin{figure}[tbh]
	\centering
	\includegraphics[width=0.99\linewidth, clip, trim=0 0 0 0]{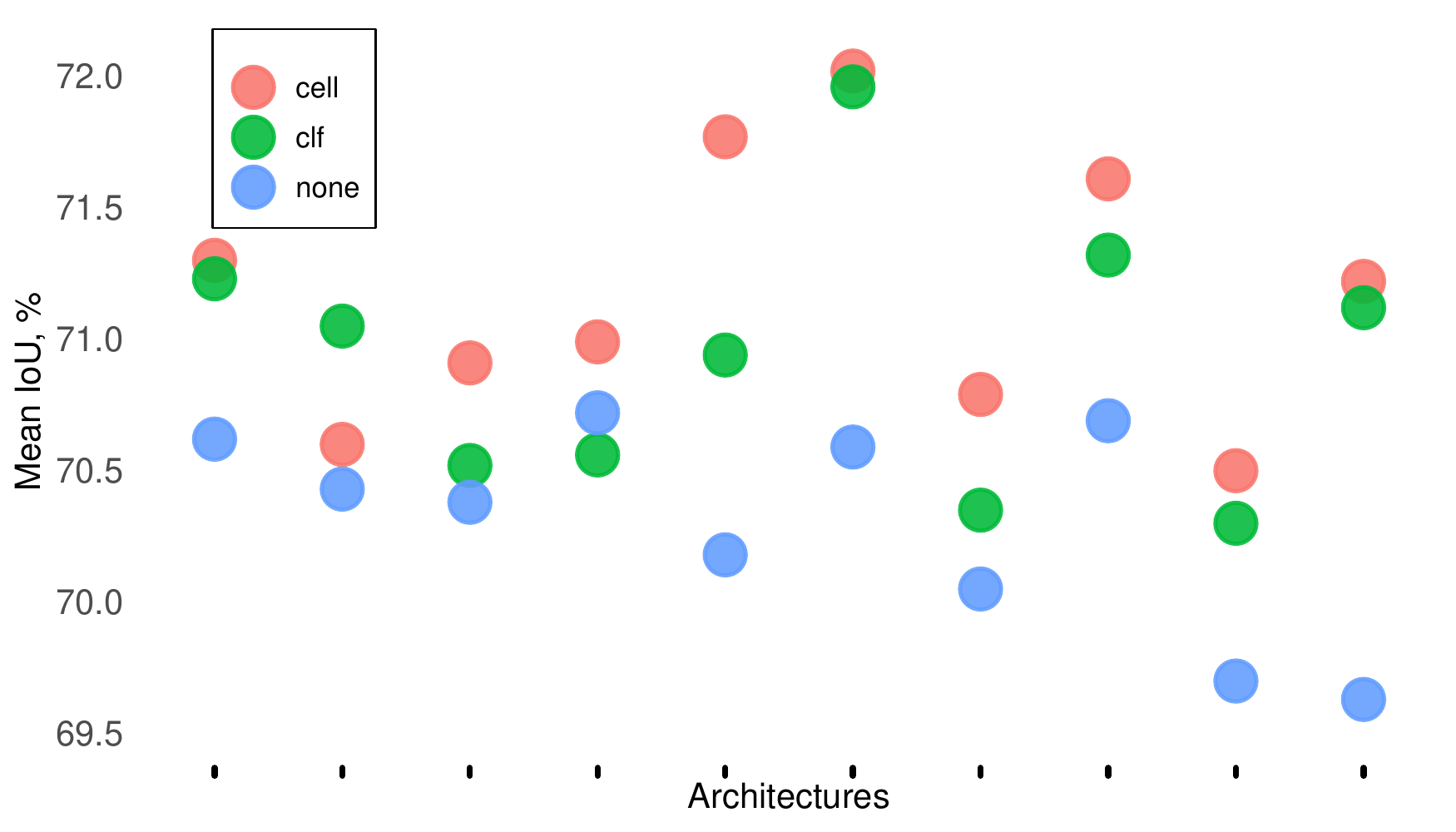}
	\vskip -0.1in
	\caption{Ablation studies on the value of intermediate supervision (\emph{none}), and the type of supervision (\emph{cell} or \emph{clf}). Each tick on the $x$-axis corresponds to a different architecture.
	\label{fig:aux-agg}}
	\vskip -0.1in
\end{figure}

The results of these ablation studies are given in Fig.~\ref{fig:aux-agg}. Auxiliary supervised architectures achieve significantly higher mean IoU, and, in particular, architectures with auxiliary cells attain best results in $8$ out of $10$ cases, reaching $3$ absolute best values across all the setups and architectures.

\subsection{Relation between search rewards and training performance}

\begin{figure}[htb]
	\subfloat[ $\rho=0.9341$\label{fig:corr-st12}]{%
		\begin{minipage}{0.5\linewidth}
			\centering
			\includegraphics[width = 1.\linewidth]{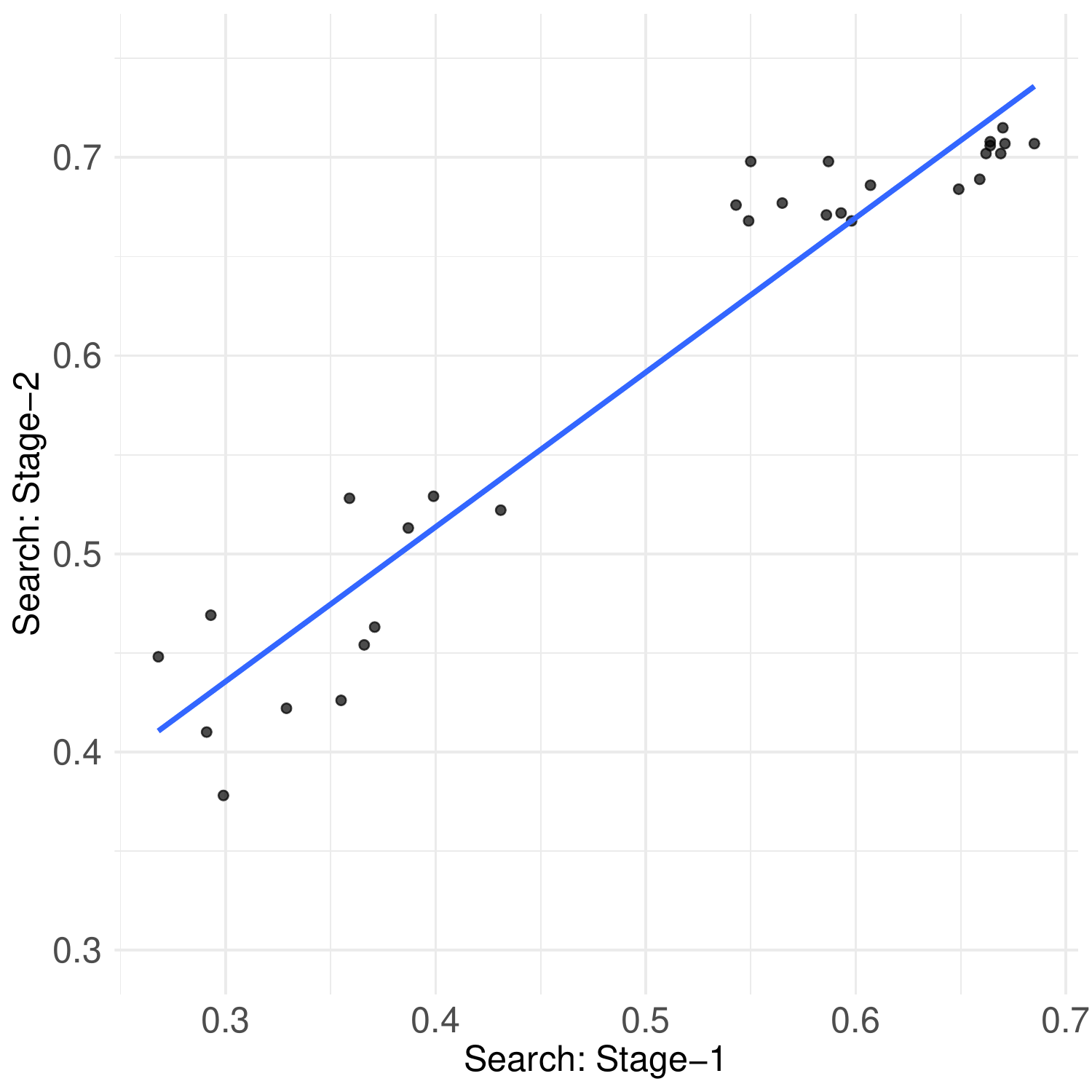}
		\end{minipage}%
	}
	\subfloat[$\rho=0.9239$\label{fig:corr-st2t}]{%
		\begin{minipage}{0.5\linewidth}
			\centering
			\includegraphics[width = 1.\linewidth,trim=0 0 0 0,clip]{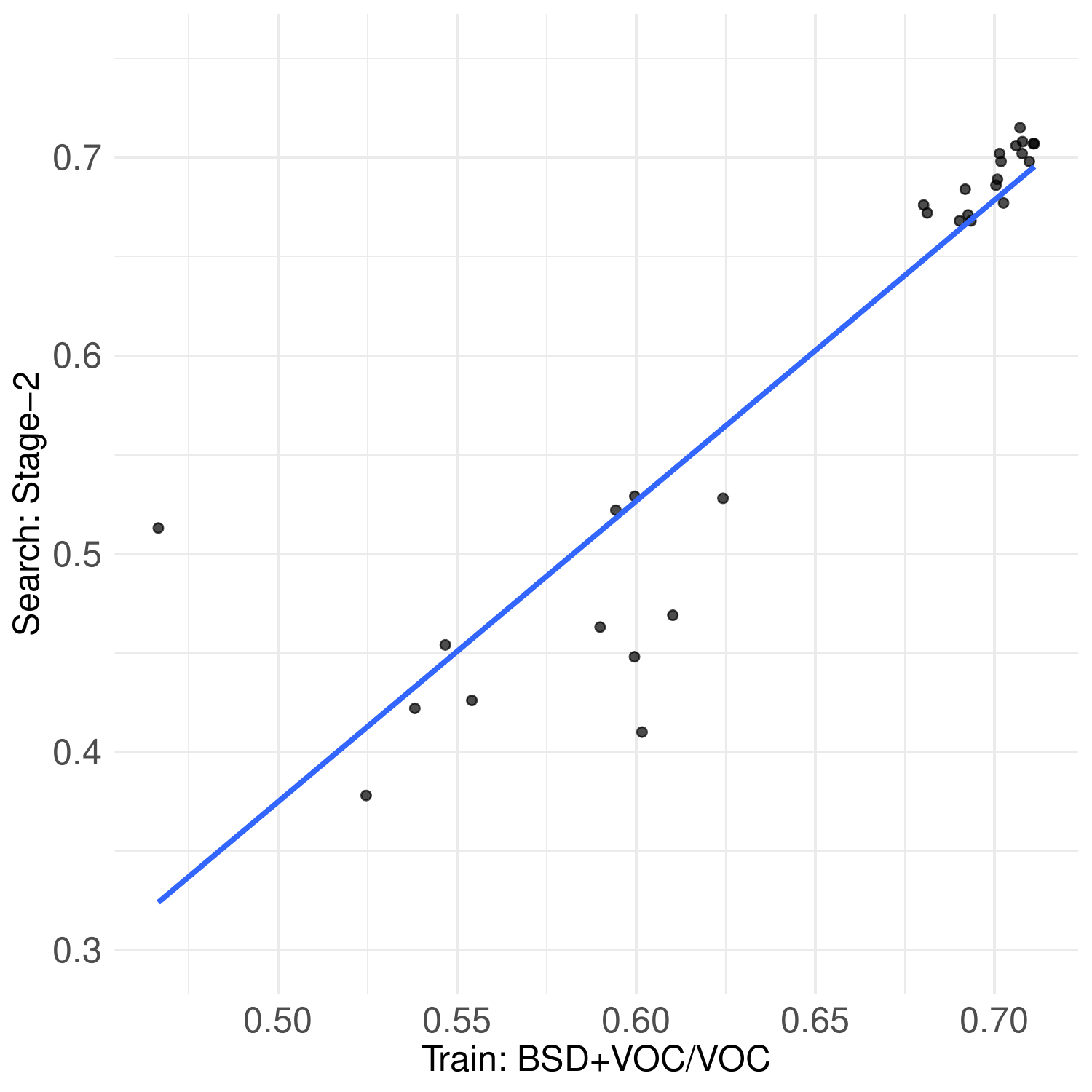}
		\end{minipage}%
	}
	\caption{Correlation between rewards acquired during search stages~\emph{\textbf{(a)}} and mean IoU after full training~\emph{\textbf{(b)}} of $30$ architectures on BSD+VOC/VOC.
	}
	\label{fig:corr}
\end{figure}

We further measure the correlation effect between rewards acquired during the search and mean IoU attained by same architectures trained for longer. To this end, we randomly sample $30$ architectures out of those explored by the controller: for fair comparison, we sample $10$ architectures with poor search performance (with rewards being less than $0.4$), $10$ with medium rewards (between $0.4$ and $0.6$), and $10$ with high rewards ($>0.6$). We train each architecture on BSD+VOC and VOC as in Sect.~\ref{subsec-eff}, rank each according to its rewards, and mean IoU, and measure the Spearman's rank correlation coefficient. As visible in Fig.~\ref{fig:corr}, there is a strong correlation between rewards after each stage, as well as between the final reward and mean IoU. This signals that our search process is able to reliably differentiate between poor-performing and well-performing architectures.

\subsection{Full Training Results}

Finally, we choose $3$ best performing architectures from Sect.~\ref{subsec-eff} and train each on the full training set, augmented with annotations from MS COCO~\cite{LinMBHPRDZ14}. The training setup is analogous to the aforementioned one with the first stage being trained for $30$ epochs (on COCO+BSD+VOC), the second stage - for $50$ (BSD+VOC), and the last one - for $100$ (VOC only). After each stage, the learning rates are halved. Additionally, halfway through the last stage we freeze the batch norm statistics and divide the learning rate in half. We exploit intermediate supervision via auxiliary cells with coefficients of $0.3, 0.25, 0.2, 0.15$ across the stages.

\begin{figure}[htb]
	\centering
	\subfloat{\includegraphics[width = 0.93\linewidth]{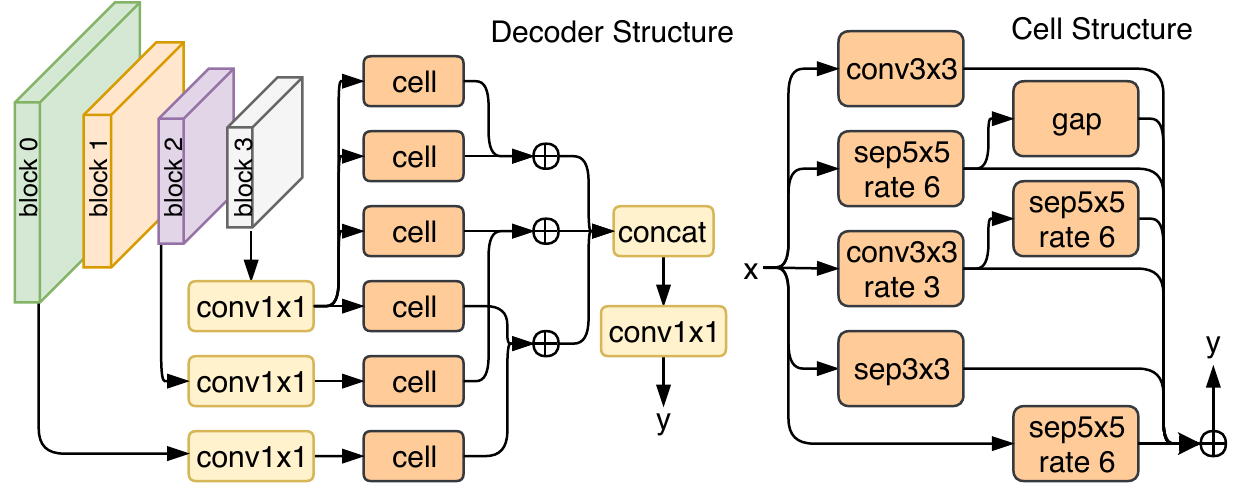}}
	\caption{Automatically discovered decoder architecture (\emph{arch0}). We visualise the connectivity structure between encoder and decoder (\emph{top}), and the cell design (\emph{bottom}). $\bigoplus$ represents an element-wise summation operation applied to each branch scaled to the highest spatial resolution among them (via bilinear interpolation), while \emph{`gap'} stands for global average pooling.
	\label{fig:v1-arch}}
	\vskip -0.16in
\end{figure}

\begin{figure}[t]
\centering
\resizebox{0.5\textwidth}{!}{\begin{tabular}{cc|ccc}
    \subfloat{\includegraphics[width = 0.19\linewidth]{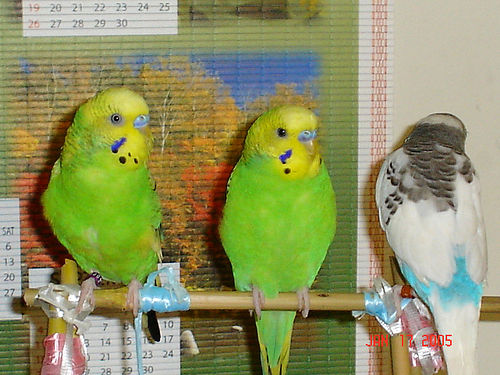}} &
    \subfloat{\includegraphics[width = 0.19\linewidth]{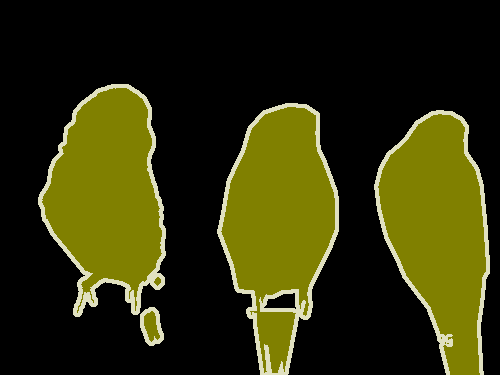}} &
    \subfloat{\includegraphics[width = 0.19\linewidth]{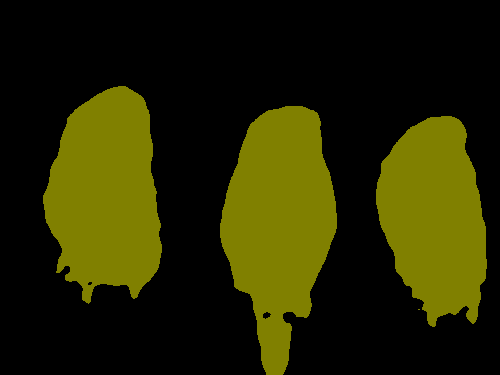}} &
    \subfloat{\includegraphics[width = 0.19\linewidth]{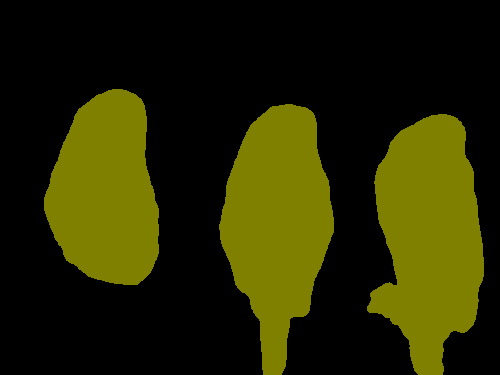}} &
    \subfloat{\includegraphics[width = 0.19\linewidth]{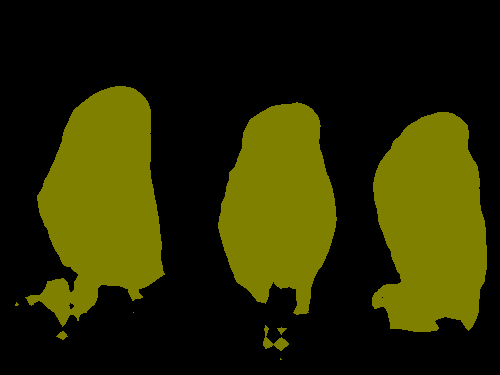}}\\[-0.15in]
    \subfloat{\includegraphics[width = 0.19\linewidth]{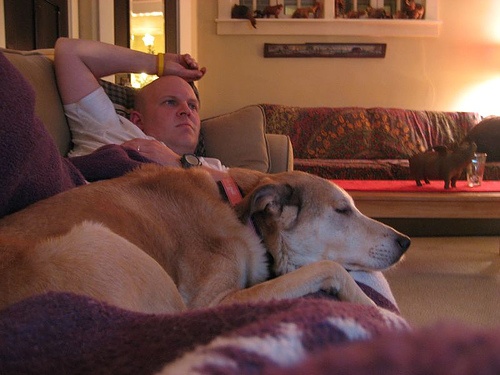}} &
    \subfloat{\includegraphics[width = 0.19\linewidth]{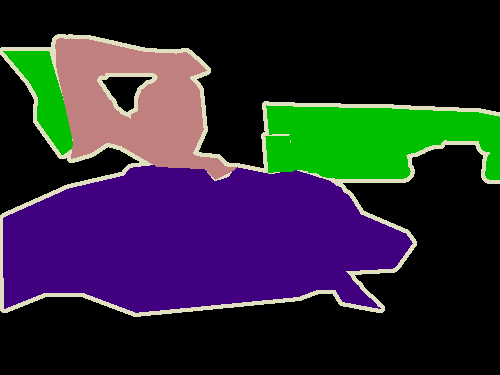}} &
    \subfloat{\includegraphics[width = 0.19\linewidth]{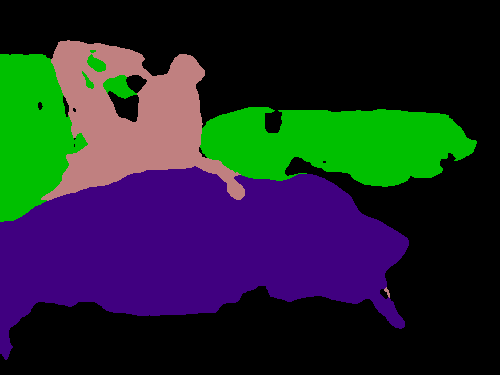}} &
    \subfloat{\includegraphics[width = 0.19\linewidth]{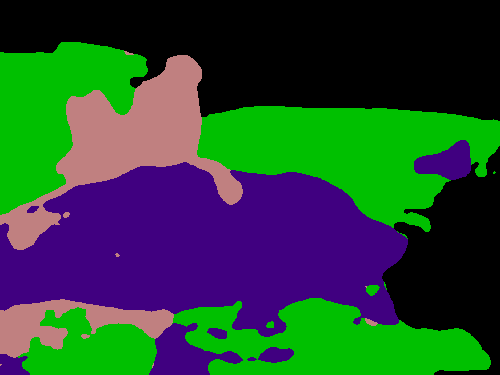}} &
    \subfloat{\includegraphics[width = 0.19\linewidth]{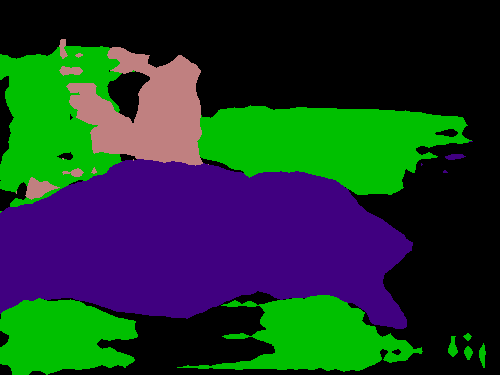}}\\[-0.15in]
    \subfloat{\includegraphics[width = 0.19\linewidth]{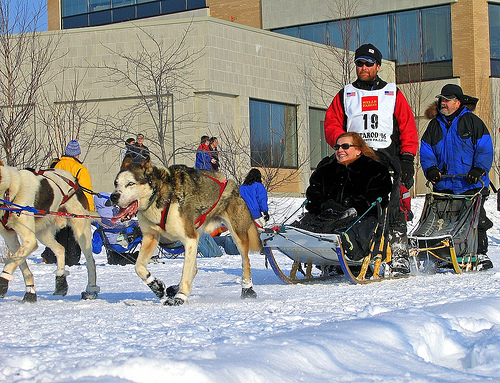}} &
    \subfloat{\includegraphics[width = 0.19\linewidth]{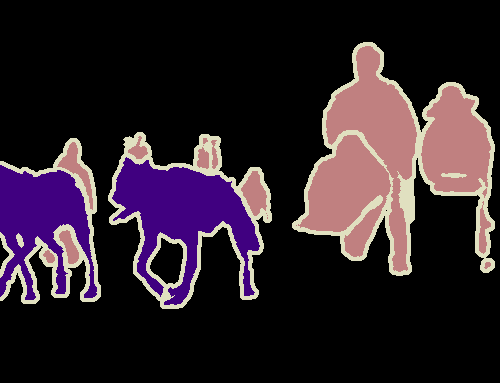}} &
    \subfloat{\includegraphics[width = 0.19\linewidth]{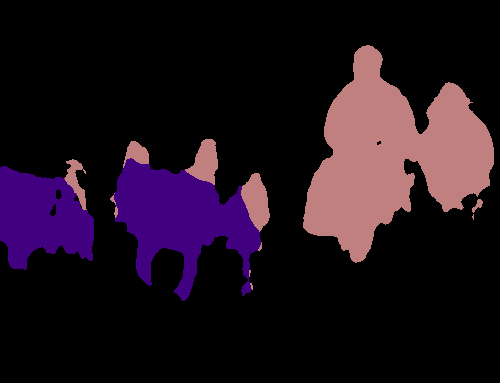}} &
    \subfloat{\includegraphics[width = 0.19\linewidth]{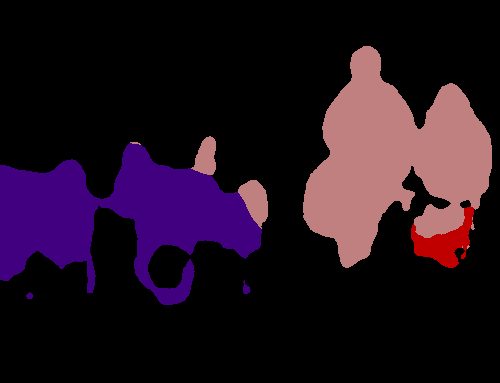}}&
    \subfloat{\includegraphics[width = 0.19\linewidth]{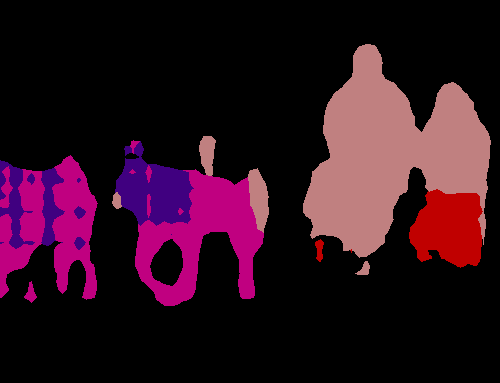}}\\[-0.15in]
    \subfloat{\includegraphics[width = 0.19\linewidth]{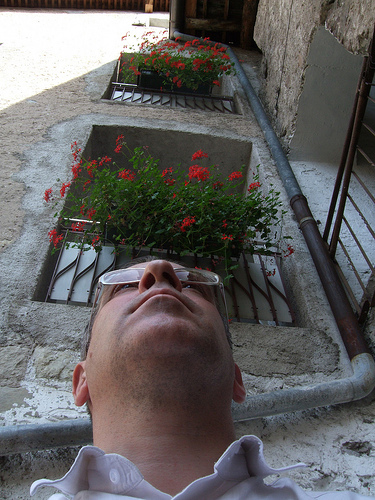}} &
    \subfloat{\includegraphics[width = 0.19\linewidth]{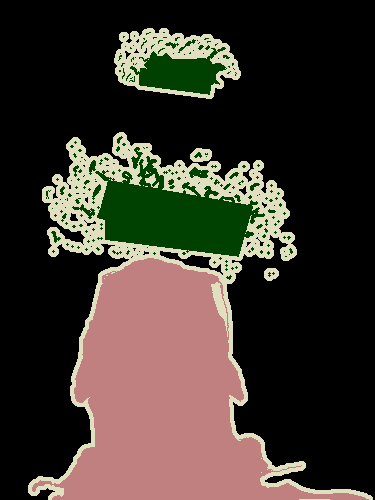}} &
    \subfloat{\includegraphics[width = 0.19\linewidth]{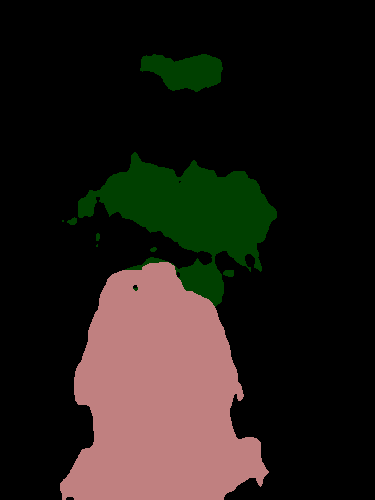}} &
    \subfloat{\includegraphics[width = 0.19\linewidth]{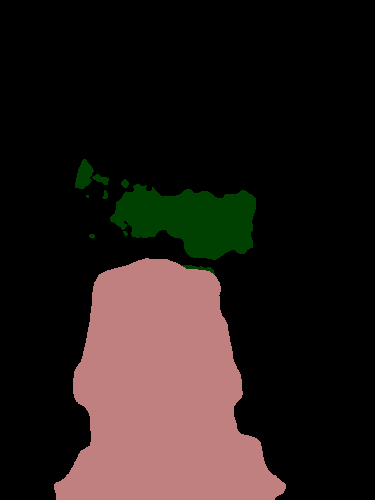}}&
    \subfloat{\includegraphics[width = 0.19\linewidth]{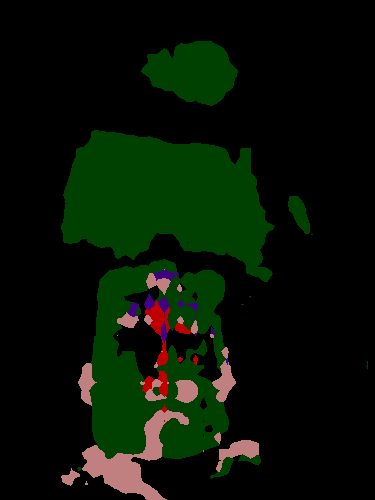}}\\[-0.15in]
    \subfloat{\includegraphics[width = 0.19\linewidth]{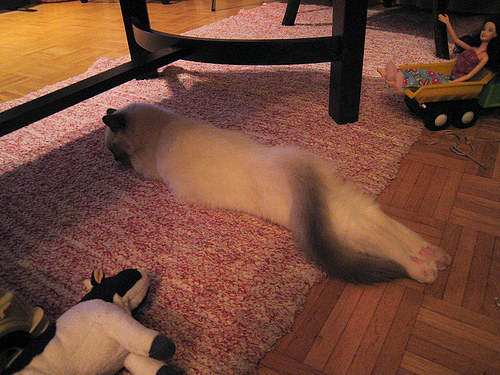}} &
    \subfloat{\includegraphics[width = 0.19\linewidth]{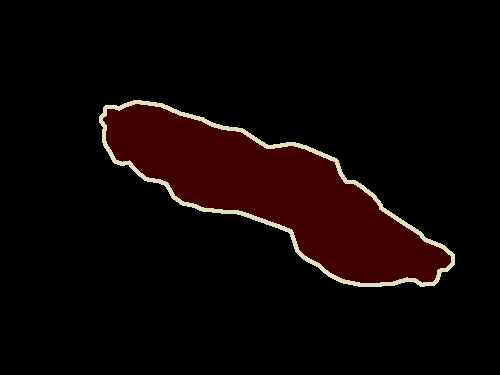}} &
    \subfloat{\includegraphics[width = 0.19\linewidth]{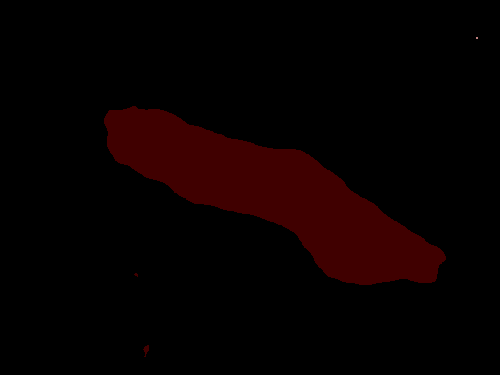}} &
    \subfloat{\includegraphics[width = 0.19\linewidth]{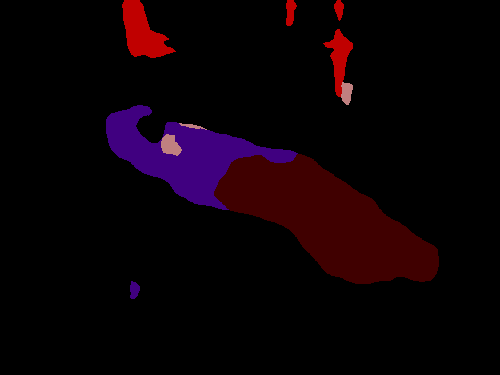}}&
    \subfloat{\includegraphics[width = 0.19\linewidth]{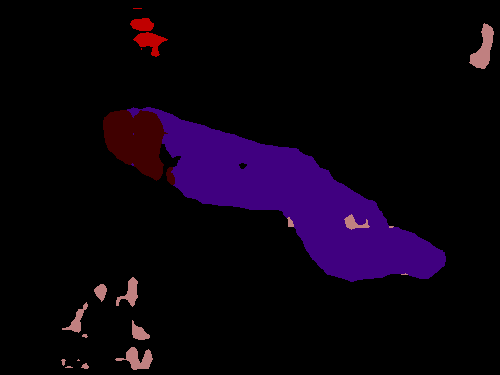}}\\
    Image&GT&Ours (arch0)&RF-LW~\cite{NekrasovS018}&DL-v3~\cite{abs-1801-04381}
\end{tabular}}
\vskip -0.1in
\caption{Inference results of our model (\emph{arch0}) on validation set of PASCAL VOC, together with Light-Weight-RefineNet (\emph{RF-LW}) and DeepLab-v3 (\emph{DL-v3}). All the models rely on MobileNet-v2 as the encoder.}
\label{fig:voc-res}
\vskip -0.1in
\end{figure}

Quantitative results are given in Table~\ref{table:full-res}.\footnote{Per-class measures are provided in {\em Appendix B}.} The architectures discovered by our method achieve competitive performance in comparison to state-of-the-art compact models and even do so with a significantly lower number of floating point operations for same output resolution. At the same time, the found architectures can be run in real-time both on a generic GPU card and JetsonTX2.\footnote{Please refer to {\em Appendix C} on notes regarding the Jetson's runtime.} Qualitatively (Fig.~\ref{fig:voc-res}), our model is able to better recognise similar and easily confused classes (\eg horse -- dog in row $3$, and cat -- dog in row $5$), better segment foreground from background and avoid spurious predictions (rows $1$,$2$,$4$,$5$).

We visualise\footnote{Other architectures are visualised in {\em Appendix A}.} the structure of the highest performing architecture (\emph{arch0}) on Fig.~\ref{fig:v1-arch}. With multiple branches encoding information of different scales, it resembles several prominent blocks in semantic segmentation, notably the ASPP module~\cite{ChenZPSA18}. Importantly, the cell found by our method differs in the way the receptive field size is controlled. Whereas ASPP solely relies on various dilation rates, here convolutions with different kernel sizes arranged in a cascaded manner allow more flexibility. Furthermore, this design is more computationally efficient and has higher expressiveness as intermediate features are easily re-used.

\subsection{Transferability to other Dense Output Tasks}

\begin{table*}[htb]
	\begin{center}
	\begin{adjustbox}{max width=0.7\textwidth}
			\begin{tabular}{c|c|c|c|c|c|c}
				\specialrule{.15em}{0em}{0em} 
				Model & Val mIoU,\%, & MAdds,B & Params,M & Output Res &\multicolumn{2}{c}{Runtime,ms (JetsonTX2/1080Ti)}\T\B\\
				\specialrule{.05em}{0em}{0em}
				\hline
				DeepLab-v3-ASPP~\cite{abs-1801-04381}&75.7&5.8&4.5&32$\times$32&69.67$\pm$0.53&\textbf{8.09$\pm$0.53}\T\\
				DeepLab-v3~\cite{abs-1801-04381}&75.9&8.73&\textbf{2.1}&64$\times$64&122.07$\pm$0.58&11.35$\pm$0.43\T\\
				RefineNet-LW~\cite{NekrasovS018}&76.2&9.3&3.3&128$\times$128&144.85 $\pm$ 0.49&12.00$\pm$0.26\T\\
				\hline
				Ours (arch0)&\textbf{78.0}&4.47&2.6&128$\times$128&109.36$\pm$0.39&14.86$\pm$0.31\T\\
				Ours (arch1)&77.1&\textbf{2.95}&2.8&64$\times$64&67.57$\pm$0.54&11.04$\pm$0.23\T\\
				Ours (arch2)&77.3&3.47&2.9&64$\times$64&\textbf{64.60$\pm$0.33}&8.86$\pm$0.26\B\\
				\specialrule{.15em}{0em}{0em}
			\end{tabular}
			\end{adjustbox}
		\caption{Results on validation set of PASCAL VOC after full training on COCO+BSD+VOC. All networks share the same backbone - MobileNet-v2. FLOPs and runtime are being measured on $512\times512$ inputs. For DeepLab-v3 we use official models provided by the authors.
			\label{table:full-res}}
	\end{center}
	\vskip -0.5in
\end{table*}

\subsubsection{Pose Estimation}
We further apply the found architectures on the task of pose estimation. In particular, the MPII~\cite{andriluka20142d} and MS COCO Keypoint~\cite{LinMBHPRDZ14} datasets are used as our benchmark. MPII includes $25$K images containing $40$K people with $16$ annotated body joints. The evaluation measure is PCKh~\cite{sapp2013modec} with thresholds of $0.5$ and $0.1$. The COCO dataset comprises $200$K images of $250$K people with $17$ body joints. Based on object keypoint similarity (OKS)\footnote{\url{http://cocodataset.org/\#keypoints-eval}}, we report average precision (AP) and average recall (AR) over $10$ different OKS thresholds.

Our quantitative results are in Table~\ref{table:pose-mpii}.\footnote{Additional qualitative and quantitative results are in {\em Appendix B}.} 
 We follow the training protocol of Xiao~\etal~\cite{xiao2018simple} and do not tune our architectures. As can be seen from the results, the discovered architectures achieve competitive performance even in comparison to a more powerful ResNet-50-based model.

\begin{table}[htb]
	\begin{center}
	    \begin{adjustbox}{max width=0.45\textwidth}
			\begin{tabular}{c|c|c|c|c|c}
				\specialrule{.15em}{0em}{0em} 
				 &\multicolumn{2}{c|}{\textbf{MPII}} & \multicolumn{2}{c|}{\textbf{COCO}} & \T\\
				 \specialrule{.1em}{0em}{0em}
				Model & Mean@0.5 & Mean@0.1 & AP & AR & Params,M \T\B\\
				\specialrule{.05em}{0em}{0em}
				\hline
				DeepLab-v3+~\cite{ChenZPSA18}&86.6&31.7&0.668&0.700&5.8\T\\
				ResNet-50~\cite{xiao2018simple}&\textbf{88.5}&\textbf{33.9}&\textbf{0.704}&\textbf{0.763}&34.0\T\\
				\hline
				Ours (arch0)&86.5&31.4&0.658&0.691&\textbf{2.6}\T\\
				Ours (arch1)&87.0&32.0&0.659&0.694&2.8\T\\
				Ours (arch2)&87.1&31.8&0.659&0.693&2.9\B\\
				\specialrule{.15em}{0em}{0em}
			\end{tabular}
		\end{adjustbox}
		\caption{Comparisons on MPII validation and COCO val2017. Flip test is used. For COCO, the same detector as in~\cite{xiao2018simple} is used for all models. DeepLab-v3+ is our re-implementation based on the official code.
			\label{table:pose-mpii}}
	\end{center}
	\vskip -0.45in
\end{table}

\iffalse
\begin{table}[htb]
	\begin{center}
	    \begin{adjustbox}{max width=0.5\textwidth}
			\begin{tabular}{c|c|c|c}
				\specialrule{.15em}{0em}{0em}
				Model & Mean & Mean@0.1 & Params,M \T\B\\
				\specialrule{.1em}{0em}{0em}
				\hline
				DeepLab-v3+~\cite{ChenZPSA18}&86.6&31.7&5.8\T\\
				ResNet-50~\cite{xiao2018simple}&\textbf{88.5}&\textbf{33.9}&34.0\T\\
				\hline
				Ours (arch0)&86.6&31.3&\textbf{2.8}\B\\
				\specialrule{.15em}{0em}{0em}
			\end{tabular}
		\end{adjustbox}
		\caption{Quantitative results on the validation set of MPII. Flip test is used. DeepLab-v3+ is our re-implementation based on the official code.
			\label{table:pose-mpii}}
			\vskip -0.35in
	\end{center}
\end{table}
\fi

\subsubsection{Depth Estimation}

Finally, we train the architectures on NYUDv2~\cite{Silberman:ECCV12} for depth prediction. Following previous work~\cite{abs-1809-04766}, we only use $25$K training images with depth annotations from the Kinect sensor, and report validation results on $654$ images in Table~\ref{table:depth-nyu}. %
 Among other compact real-time networks, we achieve significantly better results across all the metrics without any additional tricks.  
Note also that the work in \cite{abs-1809-04766} trained the depth 
model jointly with semantic segmentation, thus using extra information.
\begin{table}[htb]
    \vskip -0.11in
	\begin{center}
	\begin{adjustbox}{max width=0.45\textwidth}
			\begin{tabular}{l|c|c|c|c|c}
				\specialrule{.15em}{0em}{0em}
				&\multicolumn{3}{c|}{\textbf{Ours}}&&\T\B\\
				\specialrule{.1em}{0em}{0em}
				\hline
				 & arch0 & arch1 & arch2 & RF-LW~\cite{abs-1809-04766} & CReaM~\cite{abs-1807-08931}\T\B\\
				\specialrule{.1em}{0em}{0em}
				\hline
				RMSE (lin) & \textbf{0.523} & 0.526 & 0.525 & 0.565 & 0.687\T\B\\
				RMSE (log) & 0.184 & \textbf{0.183} & 0.189 & 0.205 & 0.251\T\B\\
				abs rel & 0.136 & \textbf{0.131} & 0.140 & 0.149 & 0.190\T\B\\
				sqr rel & 0.089 & \textbf{0.086} & 0.093 & 0.105 & $-$\T\B\\
				$\delta<1.25$ & 0.830 & \textbf{0.832} & 0.820 & 0.790 & 0.704\T\B\\
				$\delta<1.25^{2}$ & 0.967 & \textbf{0.968} & 0.966 & 0.955 & 0.917\T\B\\
				$\delta<1.25^{3}$ & \textbf{0.992} & \textbf{0.992} & \textbf{0.992} & 0.990 & 0.977\T\B \\
				\hline
				Parameters, M & 2.6 & 2.8 & 2.9 & 3.0 & \textbf{1.5}\B\\
				\specialrule{.15em}{0em}{0em}
			\end{tabular}
    \end{adjustbox}
	\caption{Quantitative results on the validation set of NYUDv2. For RMSE, abs rel and sqr rel lower values are better, whereas for accuracy ($\delta$) higher values are better.
		\label{table:depth-nyu}}
	\end{center}
	\vskip -0.55in
\end{table}

\section{Discussion and Conclusions}

There is little doubt that manual design of neural architectures is a tedious and difficult task to handle. It is even more complicated to come up with a design of compact and high-performing architecture on challenging dense prediction problems, such as semantic segmentation. In this work, we showcased a simple and reliable approach of searching for fully convolutional architectures within a reasonable amount of time and computational resources. Our method is based around over-parameterisation of small networks that allows them to converge to better solutions. We achieved competitive performance to manually designed state-of-the-art compact architectures on PASCAL VOC, while searching only for $4$ days on $2$ GPU cards. Moreover, best found architectures also attained excellent results on other dense per-pixel tasks --- pose estimation and depth prediction.

Our future goals include exploration of alternative ways of over-parameterisation and search space description.

\section*{Acknowledgements}
VN, CS, IR's participation in this work were in part supported by ARC Centre of Excellence for Robotic Vision. CS was also supported by the GeoVision CRC Project. Correspondence should be addressed to CS.

\clearpage
{\small
\bibliographystyle{ieee}
\bibliography{egbib}
}

\clearpage

\section*{Appendix A: Search Space}

\subsection*{Decoder connectivity structure}
\begin{figure}[htb]
	\centering
	\subfloat{\includegraphics[width = 0.5\linewidth]{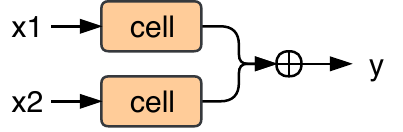}}
	\caption{Block structure of the decoder. The same cell operation is applied to two different layers specified by the connectivity configuration. If the two features have different size, the smaller one is scaled up via bilinear upsampling to match the larger one.
	\label{fig:block}}
\end{figure}

Our fully-convolutional networks follow the encoder-decoder design paradigm. In particular, in place of the encoder we rely on an existing image classifier - here, MobileNet-v2~\cite{abs-1801-04381}. The decoder has access to $4$ layers from the encoder with varying dimensions. To form connections inside the decoder part, we i.) first sample a pair of indices out of $4$ possible choices with replacement, ii.) apply the same set of operations ({\em cell}) on each sample index, iii.) sum up the outputs~(Fig.~\ref{fig:block}), and iv.) add the resultant layer into the sampling pool. In total, we repeat this process $3$ times. Finally, all non-sampled summation outputs are concatenated, before being fed into a single $1\times1$ convolution to reduce the number of channels followed by the final classification layer.

\subsection*{Cell structure}
\begin{figure}[htb]
	\centering
	\subfloat{\includegraphics[width = 0.95\linewidth]{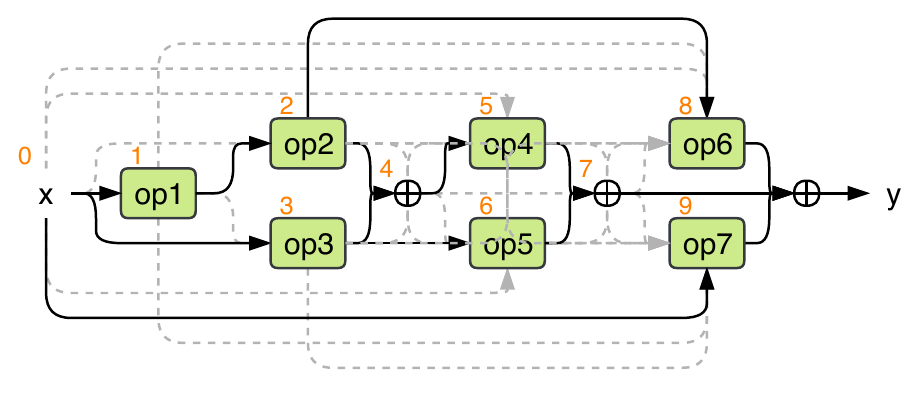}}
	\caption{Example cell structure of the decoder. The digit at the upper left corner of each operator is the index of the intermediate features. The cell is designed to utilize these features by skip connections. Except the first operator, other operators can be connected from any previous outputs. The solid black lines indicate the used paths and dashed grey lines are other unused possible paths. The cell configuration to generate the above cell is [op1, [1, 0, op2, op3], [4, 3, op4, op5], [2, 0, op6, op7]].
	\label{fig:cell}}
\end{figure}

The cell structure is similarly generated via sampling a set of operations and corresponding indices. Nevertheless, there are several notable differences:
\begin{enumerate}
    \item The operation at each position can vary;
    \item A single operation is applied to the input without any aggregation operator;
    \item After that, two indices and two operations are being sampled with replacement, with the corresponding outputs being summed up (this is repeated $3$ times);
    \item The outputs of each operation along with their summation layer are added into the sampling pool.
\end{enumerate}
An example of the cell structure with its complete search space is illustrated in Fig.~\ref{fig:cell}.

\subsection*{Architecture description}
We use a list of integers to encode the architecture found by the controller, corresponding to the output sequence of the RNN. Specifically, the list describes the connectivity structure and the cell configuration. %
For example, the following connectivity structure $[[c_1, c_2], [c_3, c_4], [c_5, c_6]]$ contains three pairs of digits, indicating the input index $c_k$ of a corresponding layer in the sampling pool. The cell configuration, $[o_1, [i_2, i_3, o_2, o_3], [i_4, i_5, o_4, o_5], [i_6, i_7, o_6, o_7]$, comprises the first operation $o_1$ followed by three cell branches with the operation $o_j$ applied on the index $i_j$. 

\begin{figure}[htb]
	\centering
	\subfloat{\includegraphics[width = 0.95\linewidth]{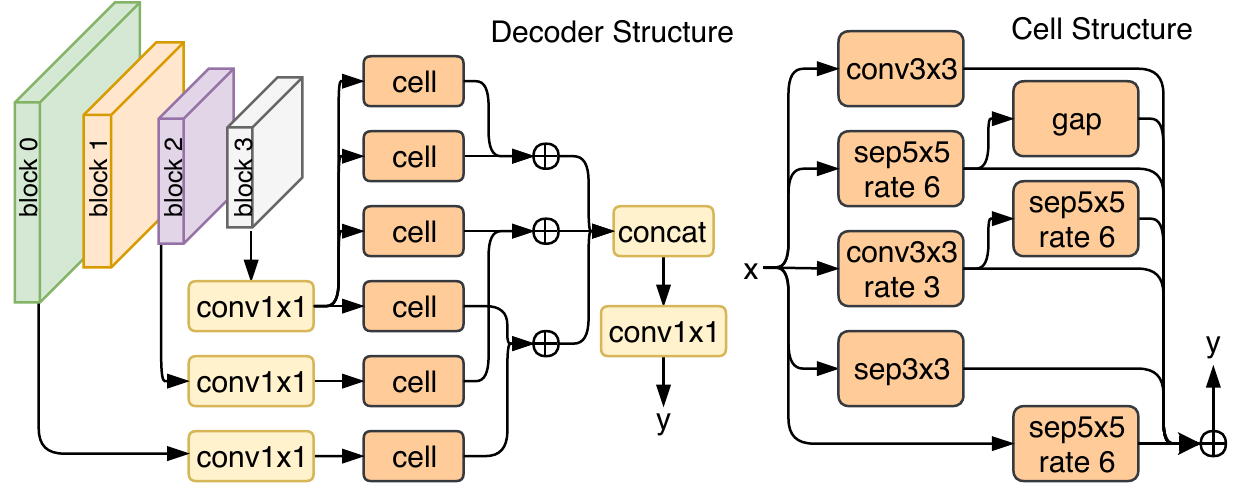}}
	\caption{arch0: $[[[3, 3], [3, 2], [3, 0]], [8, [0, 0, 5, 2], [0, 2, 8, 8], [0, 5, 1, 4]]]$
	\label{fig:arch0}}
\end{figure}

\begin{figure}[htb]
	\centering
	\subfloat{\includegraphics[width = 0.95\linewidth]{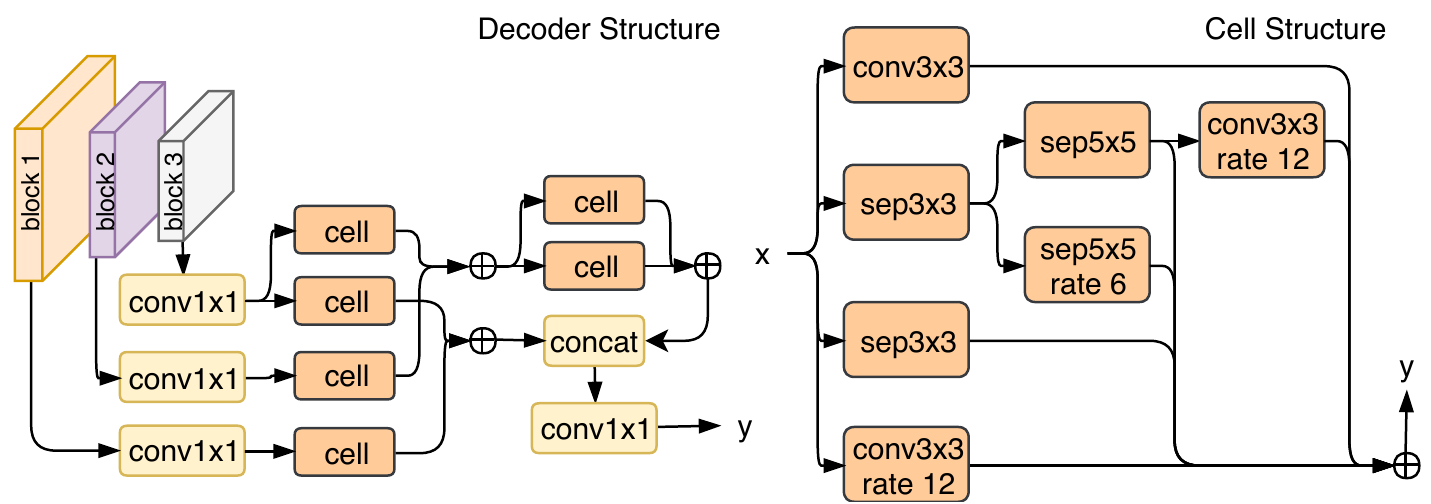}}
	\caption{arch1: $[[[2, 3], [3, 1], [4, 4]], [2, [1, 0, 3, 6], [0, 1, 2, 8], [2, 0, 6, 1]]]$
	\label{fig:arch1}}
\end{figure}

\begin{figure}[htb]
	\centering
	\subfloat{\includegraphics[width = 0.95\linewidth]{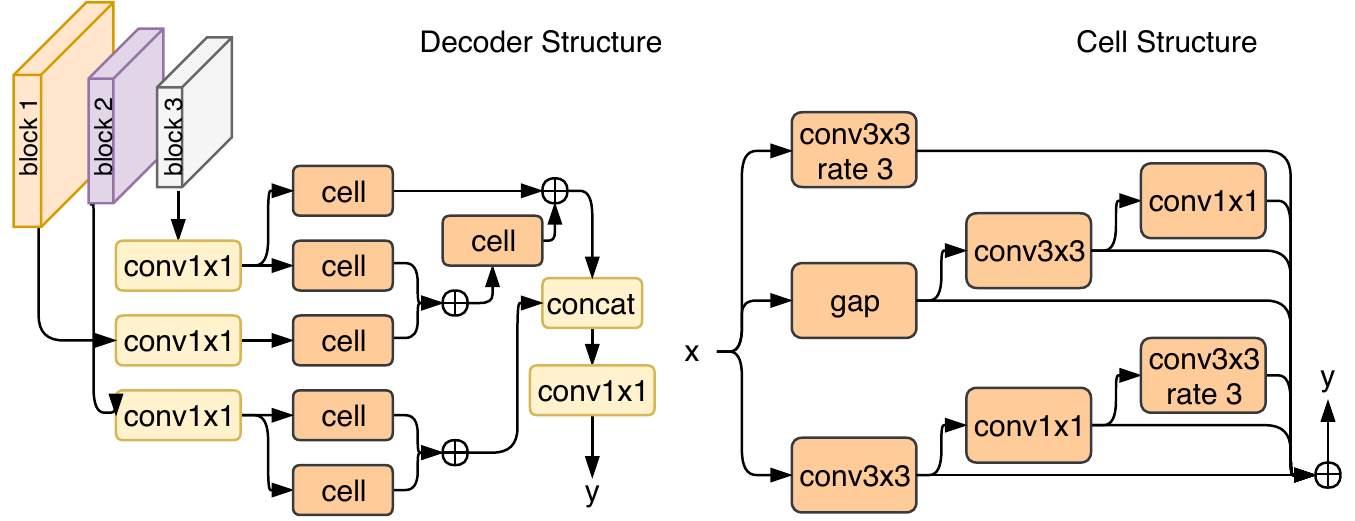}}
	\caption{arch2: $[[[1, 3], [4, 3], [2, 2]], [5, [0, 0, 4, 1], [3, 2, 0, 1], [5, 6, 5, 0]]]$
	\label{fig:arch2}}
\end{figure}

We provide the description of operations in Table~\ref{table:op-symbols}, and visualise the discovered structures in Fig.~\ref{fig:arch0} ({\em arch0}), Fig.~\ref{fig:arch1} ({\em arch1}), and Fig.~\ref{fig:arch2} ({\em arch2}). Note that inside the cell only the final summation operator is displayed as intermediate summations would lead to identical structures.

\begin{table}[htb]
\begin{center}
    \begin{tabularx}{0.5\textwidth}{c|l|X}
			\specialrule{.15em}{0em}{0em}
			Index & Abbreviation & Description \T\B\\
			\specialrule{.05em}{0em}{0em}
			\hline
			0 & conv1x1 & conv $1{\times}1$\T\\
			1 & conv3x3 & conv $3{\times}3$\T\\
			2 & sep3x3 & separable conv $3{\times}3$\T\\
			3 & sep5x5 & separable conv $5{\times}5$\T\\
			4 & gap & global average pooling followed by upsampling and conv $1{\times1}$\T\\
			5 & conv3x3 rate 3 & conv $3{\times}3$ with dilation rate $3$\T\\
			6 & conv3x3 rate 12 & conv $3{\times}3$ with dilation rate $12$\T\\
			7 & sep3x3 rate 3 & separable conv $3{\times}3$ with dilation rate $3$\T\\
			8 & sep5x5 rate 6 & separable conv $5{\times}5$ with dilation rate $6$\T\\
			9 & skip & skip-connection\T\\
			10 & zero & zero-operation that effectively nullifies the path\B\\
			\specialrule{.15em}{0em}{0em}
		\end{tabularx}
	\caption{Operation indices and abbreviations used to describe the cell configuration.
	\label{table:op-symbols}}
\end{center}
\vskip -0.4in
\end{table}

\section*{Appendix B: Experimental results}
\label{appx:b}
\subsection*{Semantic Segmentation}

\subsubsection*{PASCAL VOC}

We start training with the learning rates of $1e$-$3$ and $3e$-$3$ -- for the encoder and the decoder, respectively. The encoder weights are updated using SGD with the momentum value of $0.9$, whereas for the decoder part we rely on Adam~\cite{KingmaB14} with default parameters of $\beta_{1}{=}0.9$, $\beta_{2}{=}0.99$ and $\epsilon{=}0.001$. We exploit the batch size of $64$, evenly divided over two $1080$Ti GPU cards. Each image in the batch is randomly scaled in the range of $[0.5,2.0]$, randomly mirrored, before being randomly cropped and padded to the size of $450{\times}450$. During training, in order to calculate the loss term, we upsample the logits to the size of the target mask.

In addition to the results presented in the main text, we provide per-class intersection-over-union values across the models in Table~\ref{table:voc-val}.

\begin{table*}[htb]
	\begin{center}
	    \begin{adjustbox}{max width=1.\textwidth}
	    \begin{tabular}{c|*{21}{c}|c}
				\specialrule{.15em}{0em}{0em}
				Model & bg & aero & bike & bird & boat & bottle & bus & car & cat & chair & cow & table & dog & horse & mbike & person & plant & sheep & sofa & train & tv & mean \T\B\\
				\specialrule{.05em}{0em}{0em}
				\hline
				DeepLab-v3~\cite{abs-1801-04381} & 0.94&0.873&0.416&0.849&0.647&\textbf{0.753}&0.937&0.86&\textbf{0.904}&0.391&\textbf{0.893}&0.564&\textbf{0.847}&\textbf{0.892}&0.831&0.844&0.578&0.859&0.525&0.852&0.677 & 0.759\\
				RefineNet-LW~\cite{NekrasovS018} & 0.942&\textbf{0.895}&0.594&0.872&0.761&0.669&0.912&0.85&0.876&0.383&0.801&\textbf{0.605}&0.804&0.886&0.835&0.854&\textbf{0.603}&0.843&0.479&0.834&\textbf{0.703}&0.762 \\
				\hline
				Ours (arch0) & \textbf{0.947}&0.885&0.558&0.885&0.748&0.74&\textbf{0.944}&0.868&0.898&\textbf{0.429}&0.863&0.604&0.846&0.842&0.866&0.86&0.592&\textbf{0.869}&\textbf{0.593}&\textbf{0.875}&0.669&\textbf{0.780}\T\\
				Ours (arch1) & 0.944&0.888&\textbf{0.615}&0.866&\textbf{0.781}&0.733&0.933&0.865&0.894&0.394&0.828&0.603&0.833&0.848&0.854&0.855&0.568&0.829&0.555&0.85&0.662&0.771\T\\
				Ours (arch2) & \textbf{0.947}&0.873&0.589&\textbf{0.887}&0.753&0.75&0.943&\textbf{0.885}&0.895&0.372&0.829&0.635&0.845&0.832&\textbf{0.867}&\textbf{0.866}&0.555&0.843&0.537&0.851&0.671&0.773\B\\
				\specialrule{.15em}{0em}{0em}
			\end{tabular}
			\end{adjustbox}
		\caption{Per-class intersection-over-union on the validation set of PASCAL VOC.
		\label{table:voc-val}}
	\end{center}
\end{table*}

\subsubsection*{CityScapes}

We also evaluate whether the found decoder designs work well when coupled with other encoders -- in particular, we consider ResNet-50~\cite{HeZRS16} and use the CityScapes dataset with $2975$ training and $500$ validation images. Our training strategy is as follows: we start with the learning rates of $1e{-}2$ and $5e{-}2$ and anneal them using the `poly' schedule~\cite{ChenPKMY18} for $500$ epochs. For both parts of the network we rely on SGD with the momentum value of $0.9$, and train with the batch size of $20$. Each image in the batch is randomly scaled in the range of $[0.5,2.0]$, randomly mirrored, before being randomly cropped and padded to the size of $800{\times}800$. As for PASCAL VOC, we upsample the logits to the size of the target mask.

\begin{table}[h]
	\begin{center}
	\begin{adjustbox}{max width=0.45\textwidth}
			\begin{tabular}{c|c|c|c}
				\specialrule{.15em}{0em}{0em} 
				Model & Backbone & Val mIoU,\%, & Params,M\T\B\\
				\specialrule{.05em}{0em}{0em}
				\hline
				DeepLab-v3+\cite{ChenZPSA18}&Xception-65&\textbf{78.8}&41\T\\
				BiSeNet~\cite{YuWPGYS18}&ResNet-18&78.6&49\T\B\\
				\hline
				Ours (arch0)&ResNet-50&77.1&\textbf{24.5}\T\\
				Ours (arch1)&ResNet-50&77.3&24.7\T\\
				Ours (arch2)&ResNet-50&76.0&24.8\B\\
				\specialrule{.15em}{0em}{0em}
			\end{tabular}
			\end{adjustbox}
		\caption{Results on the validation set of CityScapes.\label{table:cs-res}}
	\end{center}
\end{table}

Validation results together with comparison to few other networks are given in Table~\ref{table:cs-res}.

\subsection*{Pose estimation}

For pose estimation, we crop the human instance with fixed aspect ratios, $1{:}1$ for MPII~\cite{andriluka20142d} and $3{:}4$ for COCO~\cite{LinMBHPRDZ14}. Following Xiao~\etal~\cite{xiao2018simple}, the bounding box is further resized such that the longer side is equal to $256$. For MPII, $\pm25\%$ scale, $\pm30$ degree rotation and random flip are used for data augmentation. The scale and rotation factors for COCO are $\pm30\%$ and $\pm40$ degrees, respectively. We generate keypoint heatmaps of output stride $4$ with  Gaussian distribution with $\sigma=2$. 
The MobileNet-v2 encoder is initialised from ImageNet. We use the Adam optimiser with the base learning rate of $1e{-}3$, and reduce it by $10$ after epochs $90$ and $120$. The training terminates at the epoch $140$. We use the batch size of $128$ evenly split between two $1080$Ti GPU cards.

\begin{figure*}[h!]
\centering
\resizebox{0.75\textwidth}{!}{\begin{tabular}{c|ccc|cc}
        \subfloat{\includegraphics[width = 0.25\linewidth]{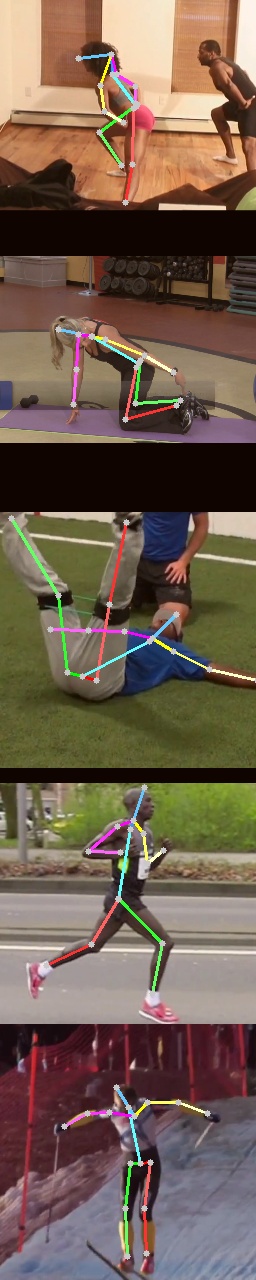}} &
        \subfloat{\includegraphics[width = 0.25\linewidth]{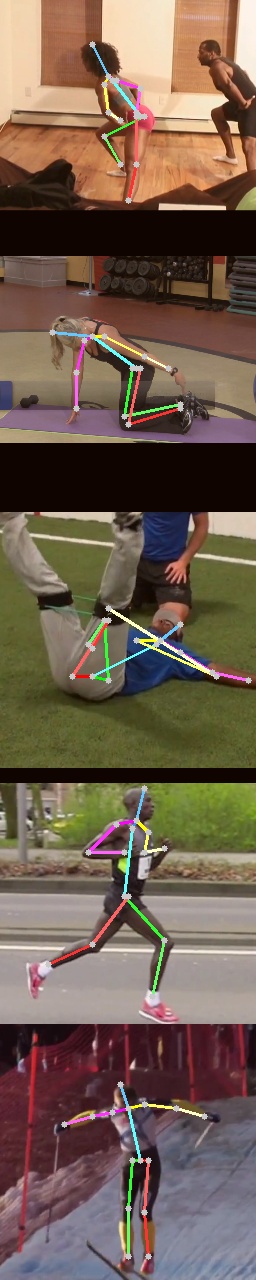}} &
        \subfloat{\includegraphics[width = 0.25\linewidth]{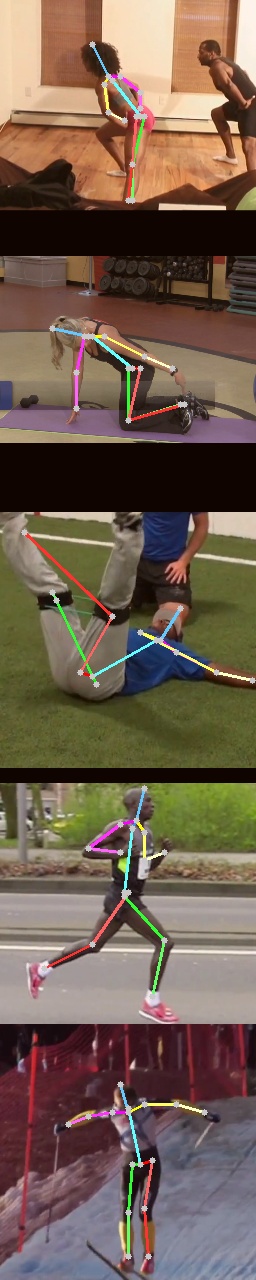}} &
        \subfloat{\includegraphics[width = 0.25\linewidth]{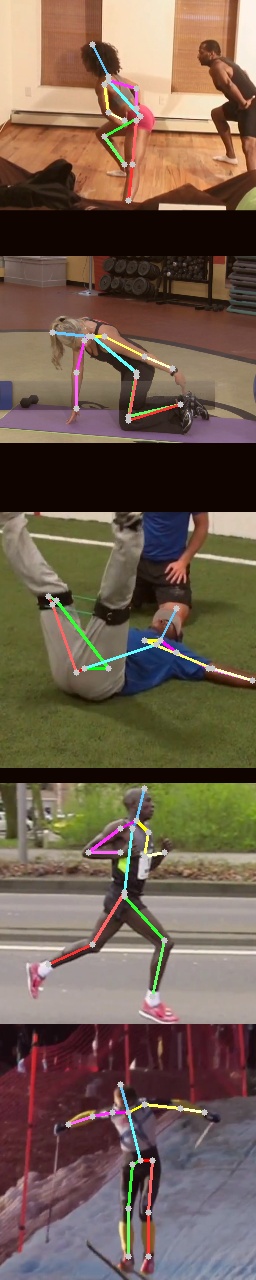}} &
        \subfloat{\includegraphics[width = 0.25\linewidth]{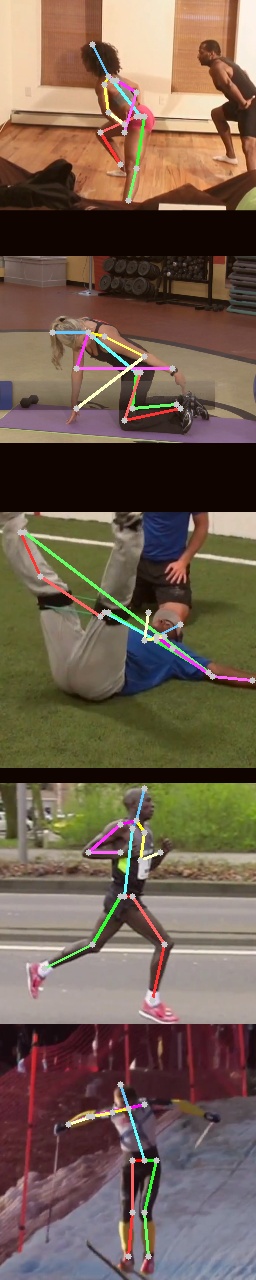}} &
        \subfloat{\includegraphics[width = 0.25\linewidth]{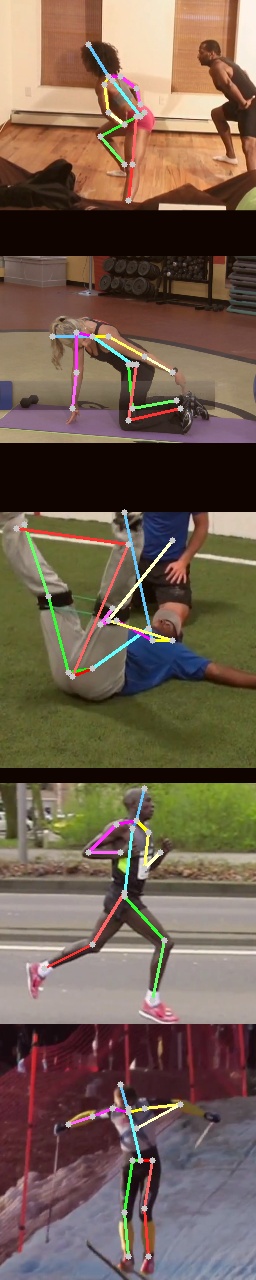}}\\[-0.15in]
        \\
        GT&arch0&arch1&arch2&DeepLab-v3+&ResNet-50~\cite{xiao2018simple}
\end{tabular}}
\caption{Inference results of our models (\emph{arch0}, \emph{arch1}, \emph{arch2}) on validation set of MPII, along with that of DeepLab-v3+-MobileNet-v2 and ResNet-50~\cite{xiao2018simple}.
\label{fig:mpii-res}}
\end{figure*}

We provide detailed quantitative results on MPII in Table~\ref{table:mpii-val} and COCO in Table~\ref{table:coco-val} along with a few qualitative examples on Fig.~\ref{fig:mpii-res}. The discovered architectures are able to infer correctly the location of the majority of keypoints (rows $1$, $2$, $4$, $5$) while failing on a more difficult input image along with other models (row $3$).

\begin{table*}[htb]
	\begin{center}
	    \begin{adjustbox}{max width=1.\textwidth}
	    \begin{tabular}{c|*{7}{c}|{c}{c}}
				\specialrule{.15em}{0em}{0em}
				Model & Head & Shoulder & Elbow & Wrist & Hip & Knee & Ankle & Mean & Mean@0.1 \T\B\\
				\specialrule{.05em}{0em}{0em}
				\hline
				DeepLab-v3+~\cite{ChenZPSA18} & 
				96.180 & 94.735 & 86.859 & 81.037 & 87.312 & 81.281 & 76.121 & 86.609 & 31.735\\
				ResNet-50~\cite{xiao2018simple} &
				\textbf{96.351} & \textbf{95.329} & \textbf{88.989} & \textbf{83.176} & \textbf{88.420} & \textbf{83.960} & \textbf{79.594} & \textbf{88.532} & \textbf{33.911} \\
				\hline
				Ours (arch0) &
				95.873 & 94.378 & 86.296 & 80.195 & 87.139 & 81.160 & 75.885 & 86.526 & 31.435\T\\
				Ours (arch1) & 
				96.317 & 94.548 & 86.501 & 80.932 & 87.242 & 81.583 & 77.374 & 86.971 & 31.951\T\\
				Ours (arch2) &
				96.146 & 94.769 & 87.097 & 80.574 & 87.848 & 81.382 & 77.586 & 87.119 & 31.782\B\\
				\specialrule{.15em}{0em}{0em}
			\end{tabular}
			\end{adjustbox}
		\caption{Per-keypoint pose estimation results on the validation set of MPII.
		\label{table:mpii-val}}
	\end{center}
\end{table*}

\begin{table*}[htb]
	\begin{center}
	    \begin{adjustbox}{max width=1.\textwidth}
	    \begin{tabular}{c|*{6}{c}}
				\specialrule{.15em}{0em}{0em}
				Model & $AP$ & $AP_{50}$ & $AP_{75}$ & $AP_m$ & $AP_l$ & $AR$ \T\B\\
				\specialrule{.05em}{0em}{0em}
				\hline
				DeepLab-v3+~\cite{ChenZPSA18} & 
				0.668&\textbf{0.894}&0.740&0.641&0.707&0.700\\
				ResNet-50~\cite{xiao2018simple} &
				\textbf{0.704}&0.886&\textbf{0.783}&\textbf{0.671}&\textbf{0.772}&\textbf{0.763} \\
				\hline
				Ours (arch0) &
				0.658&\textbf{0.894}&0.730&0.631&0.701&0.691\T\\
				Ours (arch1) & 
				0.659&0.884&0.729&0.633&0.698&0.694\T\\
				Ours (arch2) &
				0.659&0.890&0.729&0.631&0.700&0.693\B\\
				\specialrule{.15em}{0em}{0em}
			\end{tabular}
			\end{adjustbox}
		\caption{Pose estimation results on the validation set of COCO2017. We report average precision ({\em AP)} and average recall ({\em AR}). $AP_{50}$ and $AP_{75}$ stand for average precision computed with the object keypoint similarity (OKS) values of $0.5$ and $0.75$, respectively, whereas $AP_m$ and $AP_l$ are average precision metrics as measured at medium and large area ranges.
		\label{table:coco-val}}
	\end{center}
\end{table*}

\subsection*{Depth estimation}

For depth estimation, we start training with the learning rates of $1e$-$3$ and $7e$-$3$ - for the encoder and the decoder, respectively. For both we use SGD with the momentum value of $0.9$, and anneal the learning rates via the \emph{`Poly'} schedule: $lr * (1 - \frac{epoch}{400})^{0.9}$. The training is stopped after $300$ epochs. We exploit the batch size of $32$, evenly divided over two $1080$Ti GPU cards. Each image in the batch is randomly scaled in the range of $[0.5,2.0]$, randomly mirrored, before being randomly cropped and padded to the size of $500{\times}500$. We upsample the logits to the size of the target mask and use the inverse Huber loss~\cite{Laina2016} for optimisation, ignoring pixels with missing depth measurements.

\begin{figure*}[h!]
\centering
\resizebox{0.75\textwidth}{!}{\begin{tabular}{cc|ccc|c}
        \subfloat{\includegraphics[width=0.25\linewidth]{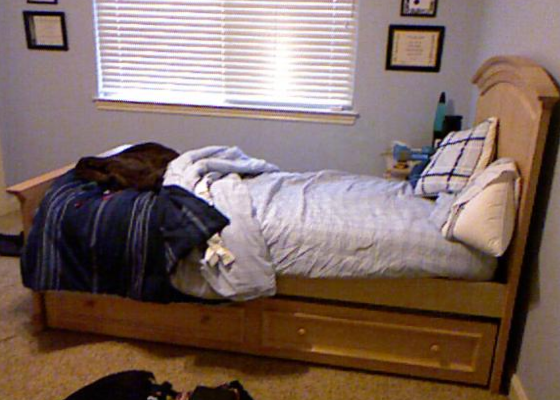}} &
        \subfloat{\includegraphics[width=0.25\linewidth]{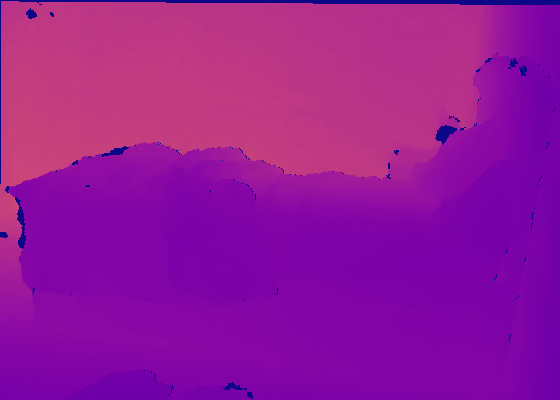}} &
        \subfloat{\includegraphics[width=0.25\linewidth]{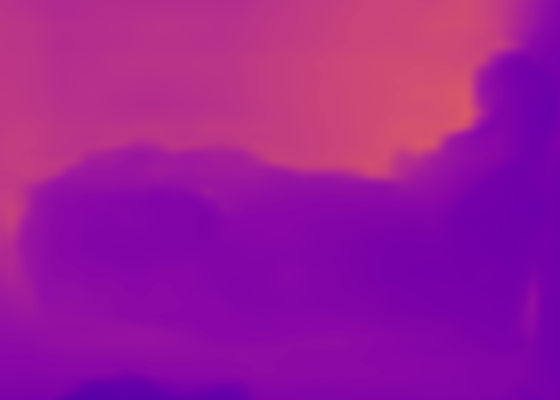}} &
        \subfloat{\includegraphics[width=0.25\linewidth]{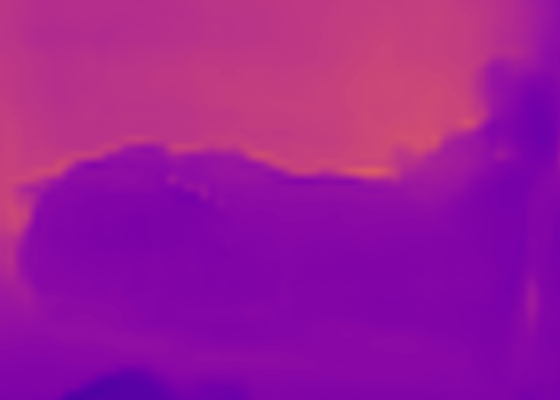}} &
        \subfloat{\includegraphics[width=0.25\linewidth]{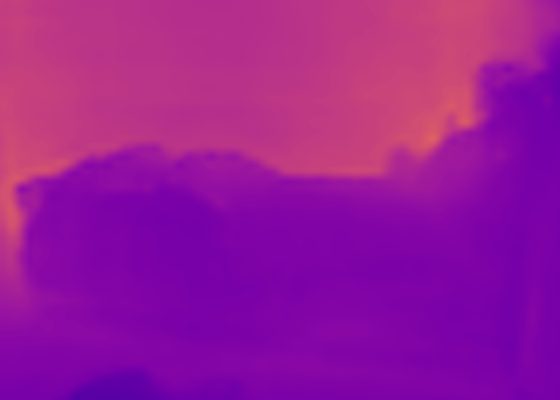}} &
        \subfloat{\includegraphics[width=0.25\linewidth]{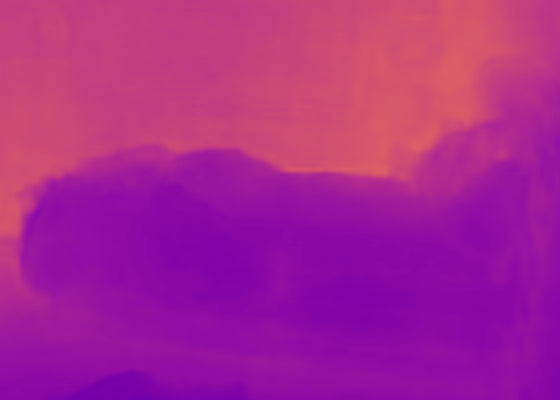}}\\[-0.15in]
        \subfloat{\includegraphics[width=0.25\linewidth]{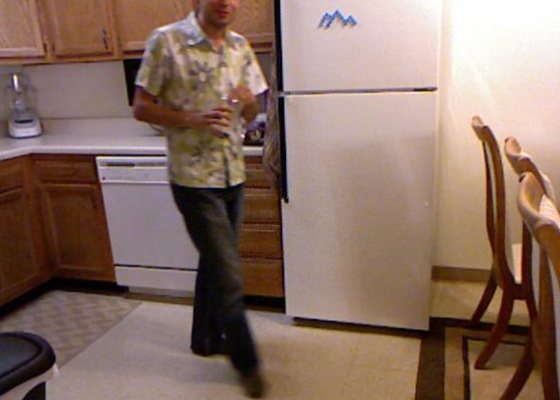}} &
        \subfloat{\includegraphics[width=0.25\linewidth]{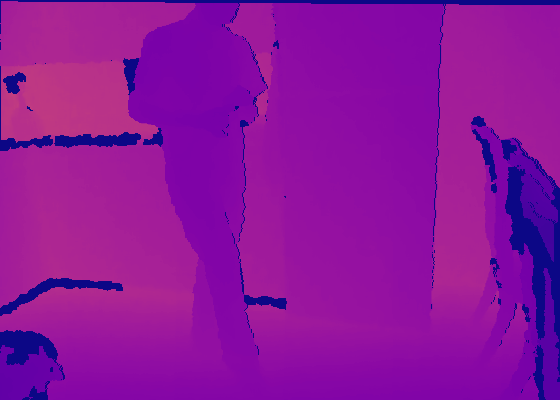}} &
        \subfloat{\includegraphics[width=0.25\linewidth]{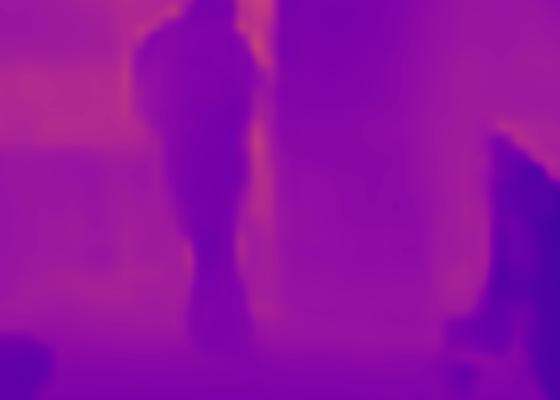}} &
        \subfloat{\includegraphics[width=0.25\linewidth]{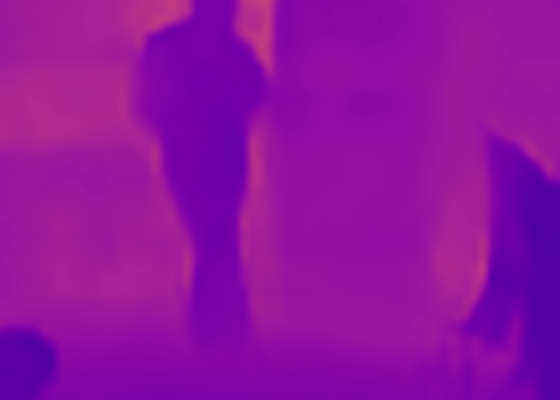}} &
        \subfloat{\includegraphics[width=0.25\linewidth]{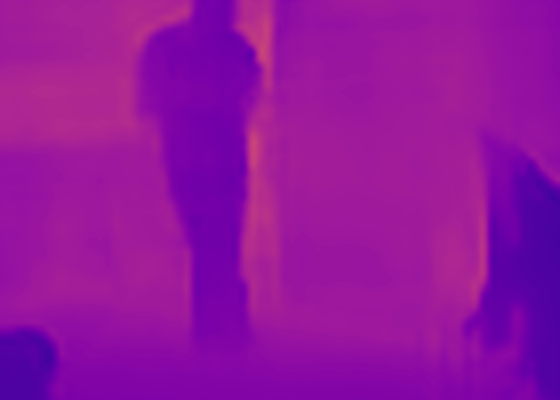}} &
        \subfloat{\includegraphics[width=0.25\linewidth]{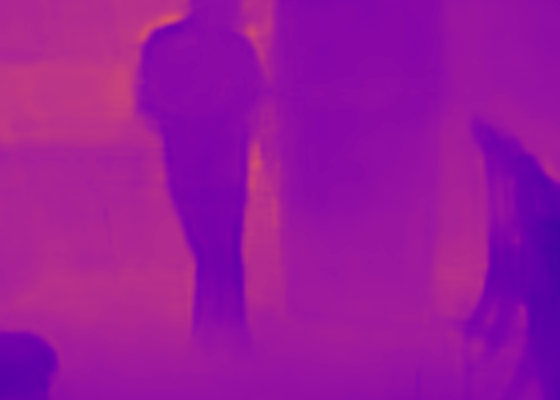}}\\[-0.15in]
        \subfloat{\includegraphics[width=0.25\linewidth]{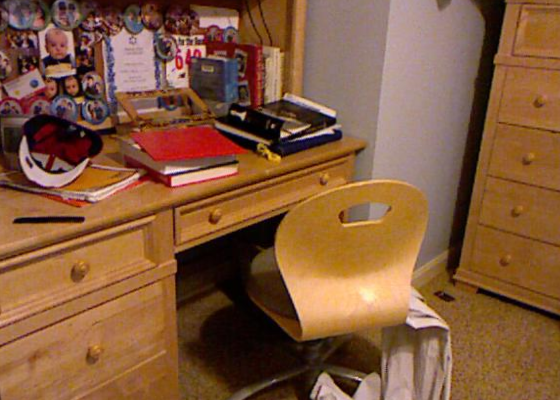}} &
        \subfloat{\includegraphics[width=0.25\linewidth]{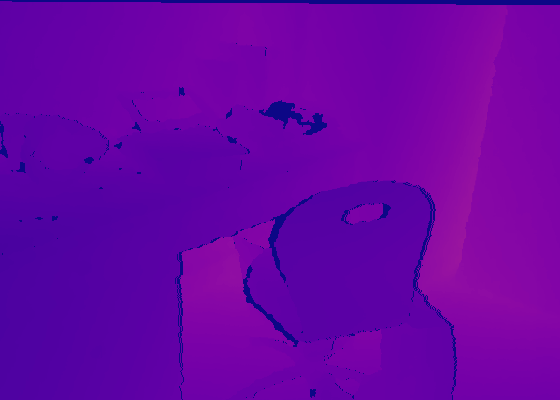}} &
        \subfloat{\includegraphics[width=0.25\linewidth]{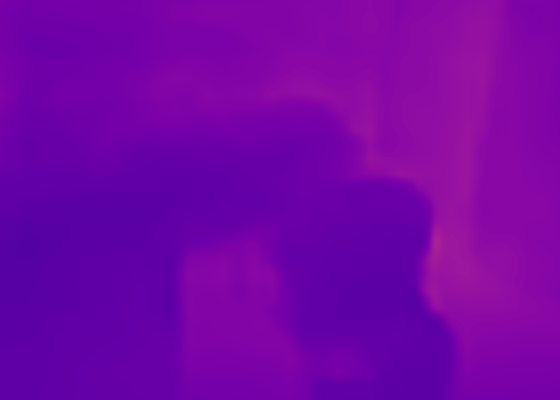}} &
        \subfloat{\includegraphics[width=0.25\linewidth]{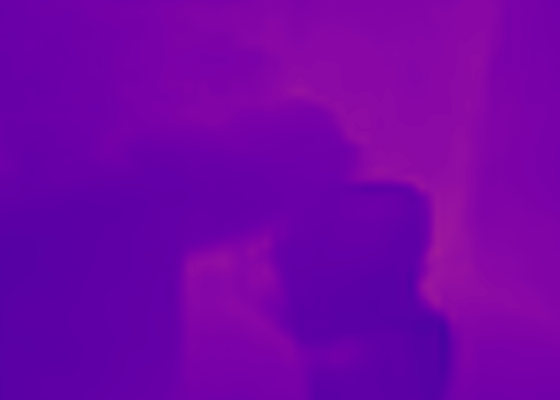}} &
        \subfloat{\includegraphics[width=0.25\linewidth]{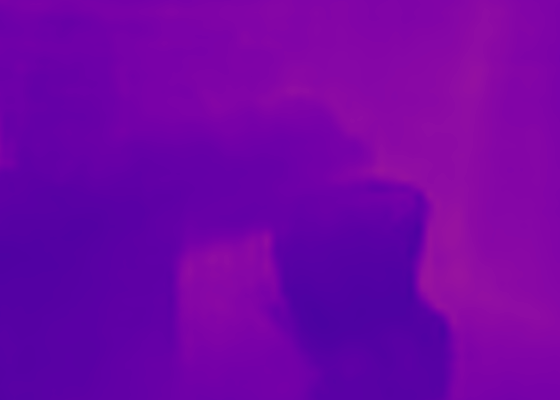}} &
        \subfloat{\includegraphics[width=0.25\linewidth]{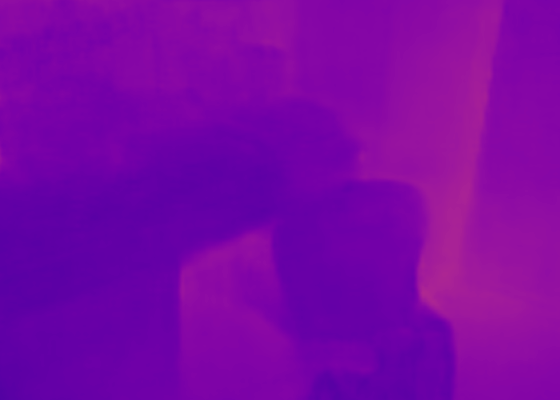}}\\[-0.15in]
        \subfloat{\includegraphics[width=0.25\linewidth]{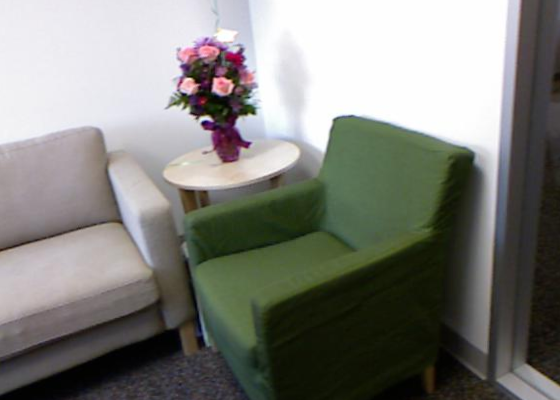}} &
        \subfloat{\includegraphics[width=0.25\linewidth]{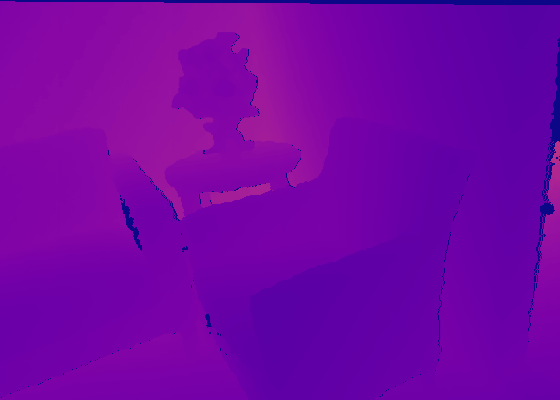}} &
        \subfloat{\includegraphics[width=0.25\linewidth]{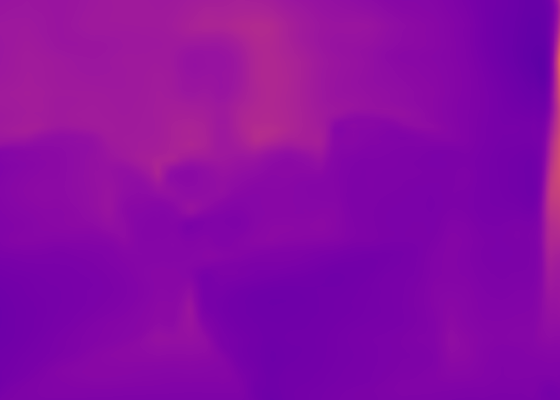}} &
        \subfloat{\includegraphics[width=0.25\linewidth]{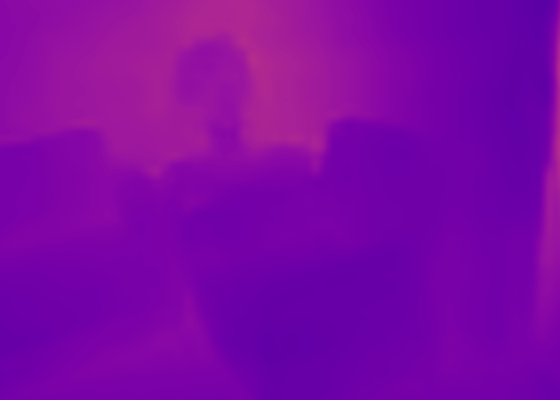}} &
        \subfloat{\includegraphics[width=0.25\linewidth]{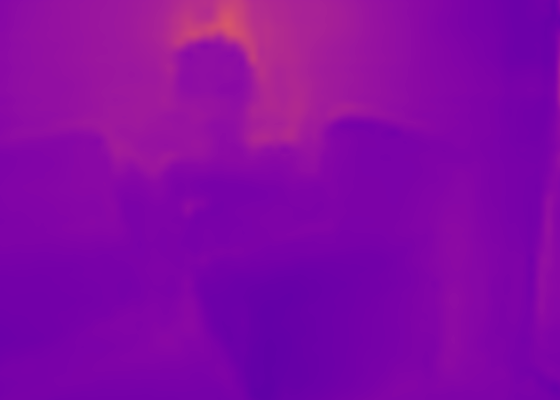}} &
        \subfloat{\includegraphics[width=0.25\linewidth]{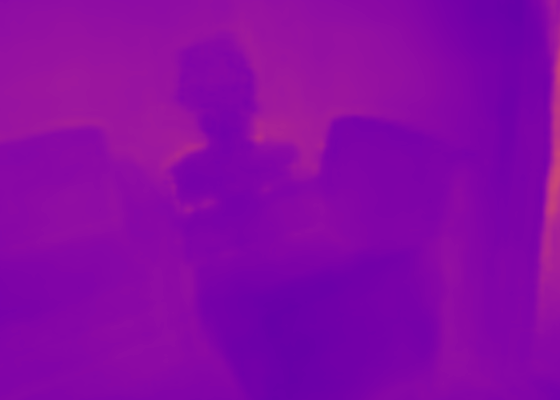}}\\
        Image&GT&arch0&arch1&arch2&RF-LW~\cite{abs-1809-04766}
\end{tabular}}
\caption{Our depth estimation qualitative results on NYUDv2, along with that of Joint Light-Weight RefineNet~\cite{abs-1809-04766}. Dark-blue pixels in ground truth are pixels with missing depth measurements.
\label{fig:nyud-res}}
\end{figure*}

We visualise qualitative results on the validation set in Fig.~\ref{fig:nyud-res}.

\section*{Appendix C: JetsonTX2 runtime}

During our experiments we observed a significant difference between models' runtime on JetsonTX2 and $1080$Ti. To better understand it, we additionally measured runtime of each discovered architecture together with Light-Weight RefineNet~\cite{NekrasovS018} varying the input resolution. 

\begin{figure*}[htbp]
\subfloat[JetsonTX2]{\includegraphics[width = 0.5\linewidth]{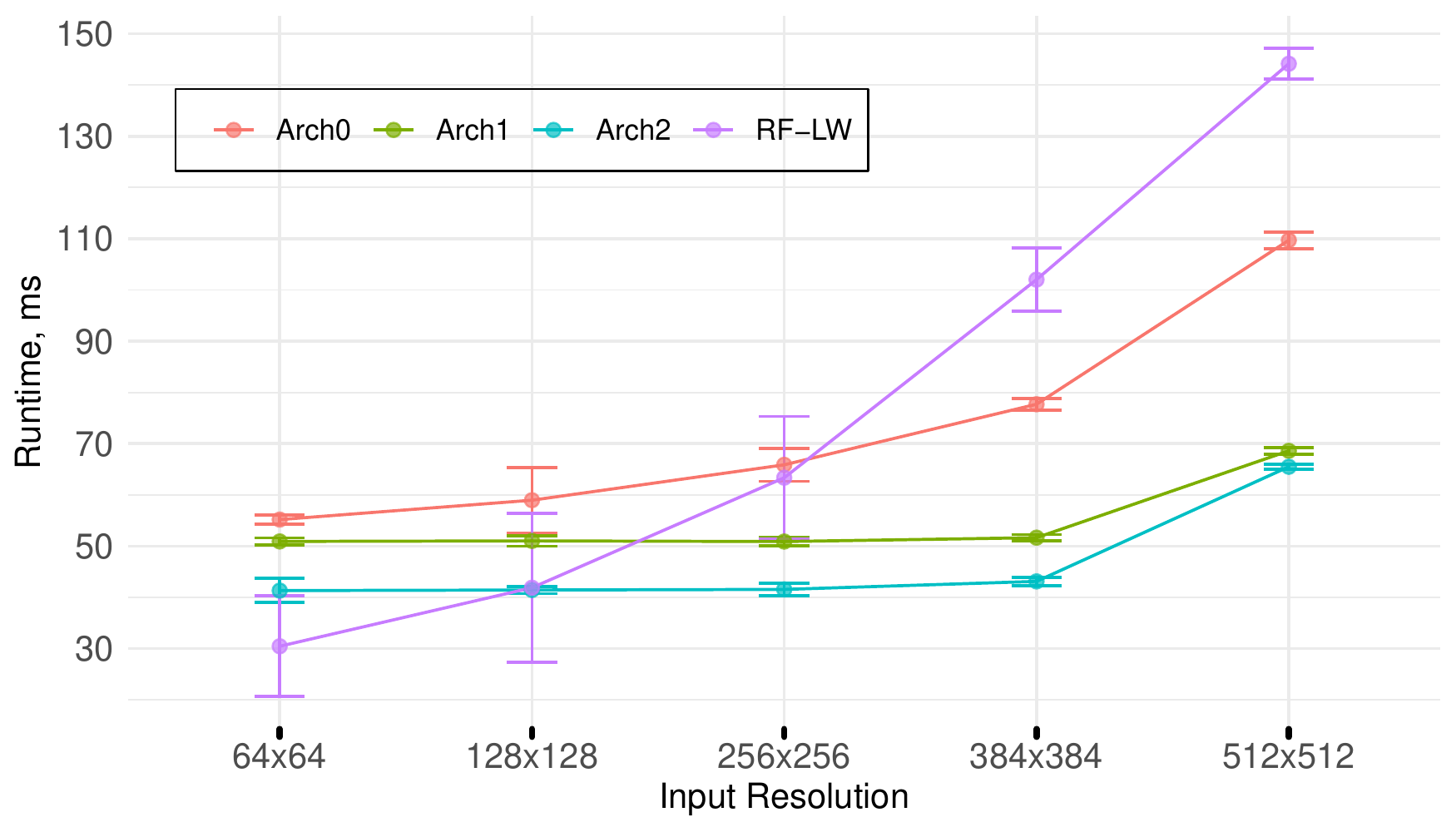}}
\subfloat[1080Ti]{\includegraphics[width = 0.5\linewidth]{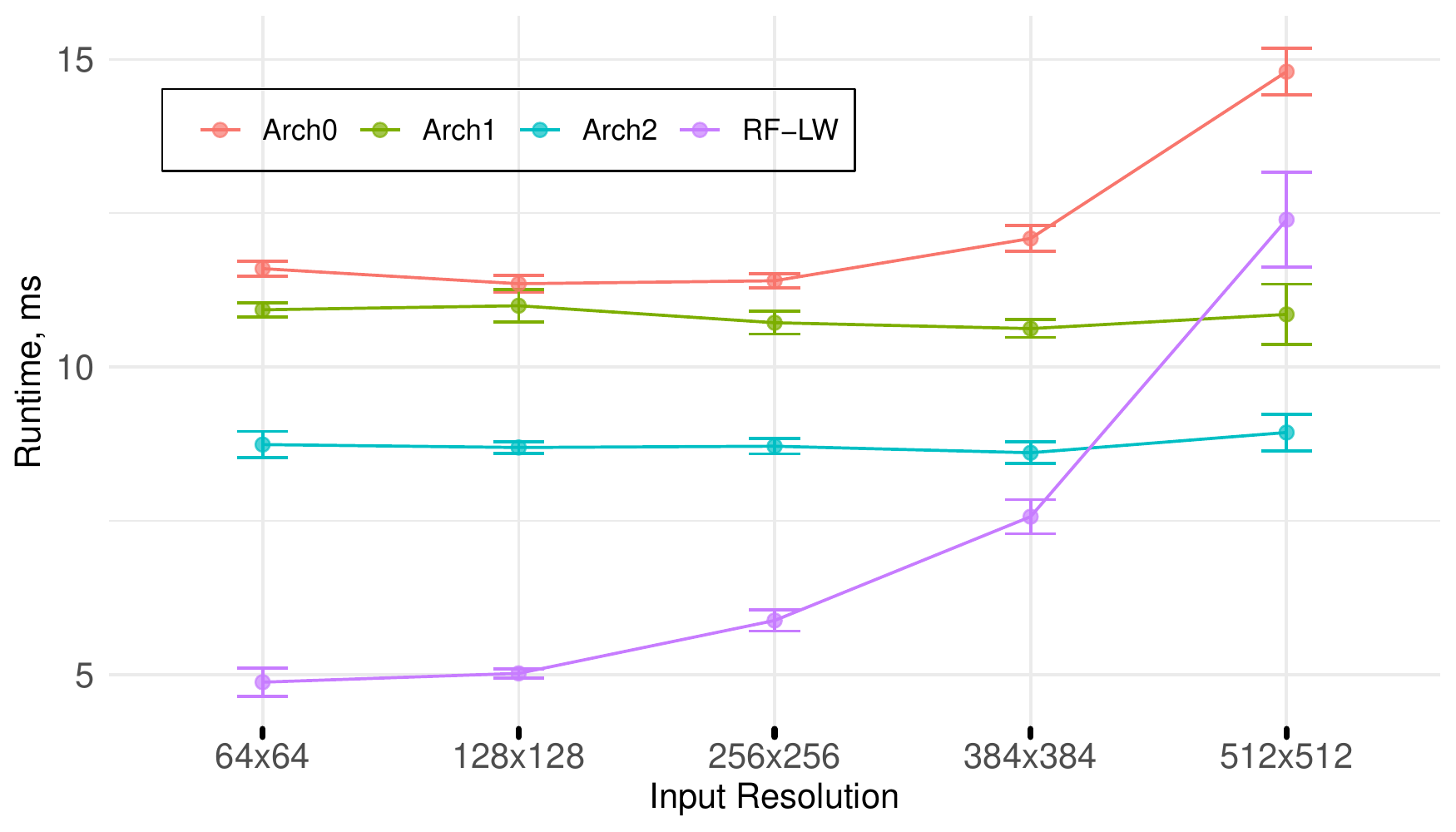}}
\caption{Models' runtime on JetsonTX2 (\textbf{a}) and 1080Ti~(\textbf{b}). We visualise mean together with standard deviation values over $100$ passes of each model.\label{fig:jetson}}
\end{figure*}

As evident from Fig.~\ref{fig:jetson}, the models with a larger number of floating point operations (i.e., {\em Arch0} and {\em RF-LW}) do not scale well with the input resolution. The effect is even more pronounced on JetsonTX2, as been independently confirmed by an NVIDIA employer in a private conversation. 

\end{document}